\documentclass[journal]{IEEEtran}

\usepackage{cite}

\ifCLASSINFOpdf
  \usepackage[pdftex]{graphicx}
  \graphicspath{{./figures/}}

\else

\fi

\usepackage{subcaption}

\usepackage{float}
\usepackage{amsmath}
\usepackage{amssymb}
\usepackage{amstext}
\usepackage{amsfonts}
\usepackage{url}
\usepackage{bbm}
\usepackage{color}

\usepackage{tablefootnote}

\usepackage{float}

\usepackage{color}
\definecolor{red}{RGB}{255,0,0}

\usepackage{multirow}

\newcommand{\revise}[1]{\textcolor{black}{#1}}

\usepackage{xspace}
\newcommand{\ie}{\textit{i.e.}\xspace}
\newcommand{\eg}{\textit{e.g.}\xspace}

\usepackage[linesnumbered,lined,noend,ruled]{algorithm2e}

\begin{document}
%
\title{TextField: Learning A Deep Direction Field for Irregular Scene Text Detection}

\author{Yongchao Xu,
        \and Yukang Wang,
        \and Wei Zhou,
        \and Yongpan Wang,
        \and Zhibo Yang,
        \and Xiang Bai, \textit{Senior Member, IEEE}
    \thanks{Yongchao Xu, Yukang Wang, Wei Zhou, and Xiang Bai are with the School of Electronic Information and Communications, Huazhong University of Science and Technology (HUST), Wuhan, 430074, China. Yongpan Wang and Zhibo Yang are with Alibaba Group. Email: \{yongchaoxu, wangyk, weizhou, xbai\}@hust.edu.cn, \and yongpan@taobao.com, \and zhibo.yzb@alibaba-inc.com. \textit{(Corresponding
author: Xiang Bai.)}}
}




\maketitle

\begin{abstract}

Scene text detection is an important step of scene text reading system. The main challenges lie on significantly varied sizes and aspect ratios, arbitrary orientations and shapes. Driven by recent progress in deep learning, impressive performances have been achieved for multi-oriented text detection. Yet, the performance drops dramatically in detecting curved texts due to the limited text representation (\eg, horizontal bounding boxes, rotated rectangles, or quadrilaterals). It is of great interest to detect curved texts, which are actually very common in natural scenes. In this paper, we present a novel text detector named TextField for detecting irregular scene texts. Specifically, we learn a direction field pointing away from the nearest text boundary to each text point. This direction field is represented by an image of two-dimensional vectors and learned via a fully convolutional neural network. It encodes both binary text mask and direction information used to separate adjacent text instances, which is challenging for classical segmentation-based approaches. Based on the learned direction field, we apply a simple yet effective morphological-based post-processing to achieve the final detection. Experimental results show that the proposed TextField outperforms the state-of-the-art methods by a large margin (28\% and 8\%) on two curved text datasets: Total-Text and SCUT-CTW1500, respectively, and also achieves very competitive performance on multi-oriented datasets: ICDAR 2015 and MSRA-TD500. Furthermore, TextField is robust in generalizing to unseen datasets.
The code is available at \url{https://github.com/YukangWang/TextField}.

\end{abstract}

\IEEEpeerreviewmaketitle

\section{Introduction}
\label{sec:introduction}
Scene text frequently appears on many scenes and carries important information for many applications, such as product search~\cite{xiong2016text}, scene understanding~\cite{yi2014scene,kang2017detection}, and autonomous driving~\cite{rong2016recognizing,zhu2018cascaded}. Scene text reading is thus of great importance. As compared to general object detection, scene text detection, the prerequisite step of scene text recognition, faces particular challenges~\cite{ye2015text} due to significantly varied aspect ratios and sizes (usually small), uncontrollable lighting conditions, arbitrary orientations and shapes. To cope with these challenges, traditional methods~\cite{wang2010word, pan2011hybrid,neumann2012real, bai2013scene,huang2013text,huang2014robust,yin2015multi,lu2015scene,tian2015text} tend to involve complete pipelines and resort to specifically engineered features. The traditional pipeline usually consists of candidate character/word generation~\cite{matas2004robust,epshtein2010detecting}, candidate filtering~\cite{jaderberg2016reading} and grouping~\cite{bai2013scene,yin2015multi}. Each module requires careful parameter tuning and specifical heuristic rules designing to make it work properly. It is thus difficult to optimize the whole pipeline, and also results in low detection speed.

Thanks to recent development of object detection~\cite{ren2017faster, liu2016ssd,redmon2016you} and segmentation~\cite{long2015fully} with deep learning, scene text detection has witnessed a great progress~\cite{liu2017deep,liao2017textboxes,zhou2017east,he2017single,he2017DDR,hu2017wordsup,ma2018arbitrary,liao2018textboxes++,liao2018rotation,wang2018geometry,tian2016ctpn,shi2017seglink,lyu2018multi,liu2018learning,zhang2016textfcn,yao2016scene,wu2017self,he2017multi,ch2017total,tian2017wetext,xue2018accurate,he2018end,zhan2018verisimilar}. They can be roughly divided into three categories: 1) Regression-based methods. Scene text is a specific type of object. Many recent
methods~\cite{liu2017deep,liao2017textboxes,he2017single,hu2017wordsup,ma2018arbitrary,liao2018textboxes++,liao2018rotation,wang2018geometry} adapt the general object detection framework to detect texts by directly regressing horizontal/oriented rectangles or quadrilaterals, which enclose texts. Some other methods attempt to regress text parts~\cite{tian2016ctpn,shi2017seglink,liu2018learning,tian2017wetext} or corners~\cite{lyu2018multi} followed by a linking or combination process. 2) Segmentation-based methods. Scene text detection can also be regarded as text instance segmentation. Several methods~\cite{zhang2016textfcn,yao2016scene,wu2017self,he2017multi,ch2017total} rely on fully convolutional network to segment text areas. A heavy post-processing is usually involved to extract text instances from the segmented text areas. 3) Hybrid methods. Some other methods~\cite{zhou2017east,he2017DDR} predict text score maps via segmentation and then obtain bounding boxes via regression.

\begin{figure}
\centering

\begin{subfigure}[b]{0.32\linewidth}
\centering
\includegraphics[width=1.0\linewidth]{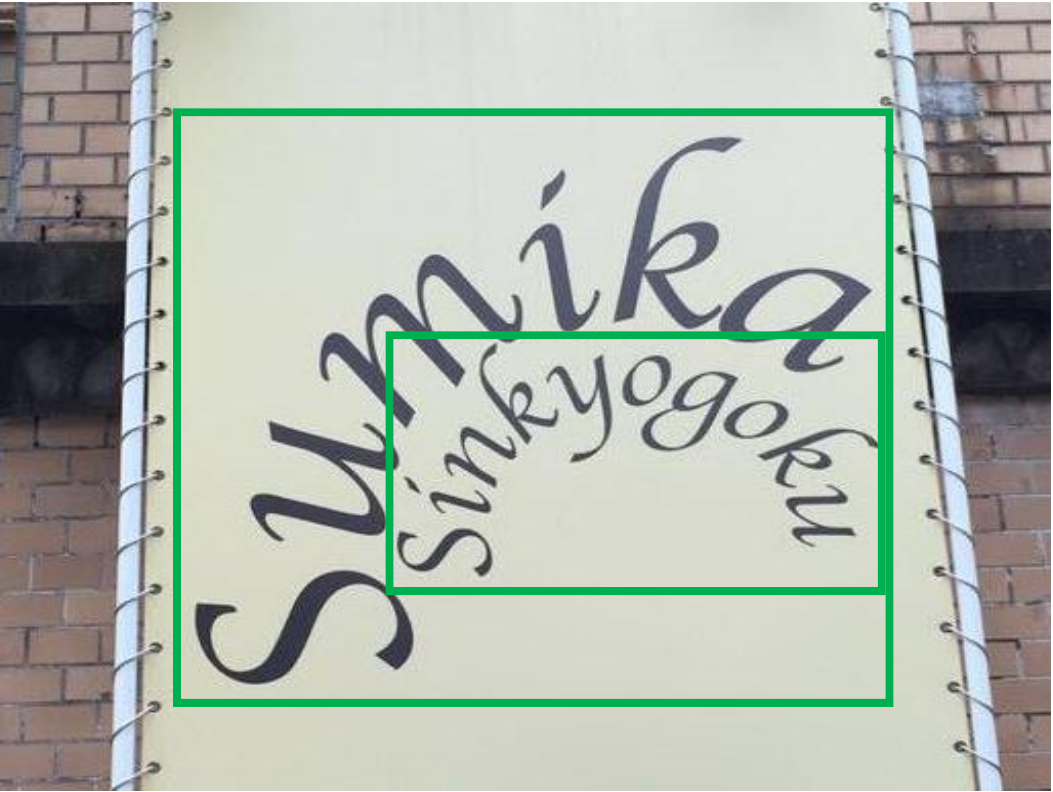}
\caption{Horizontal box}
\end{subfigure}
\begin{subfigure}[b]{0.32\linewidth}
\centering
\includegraphics[width=1.0\linewidth]{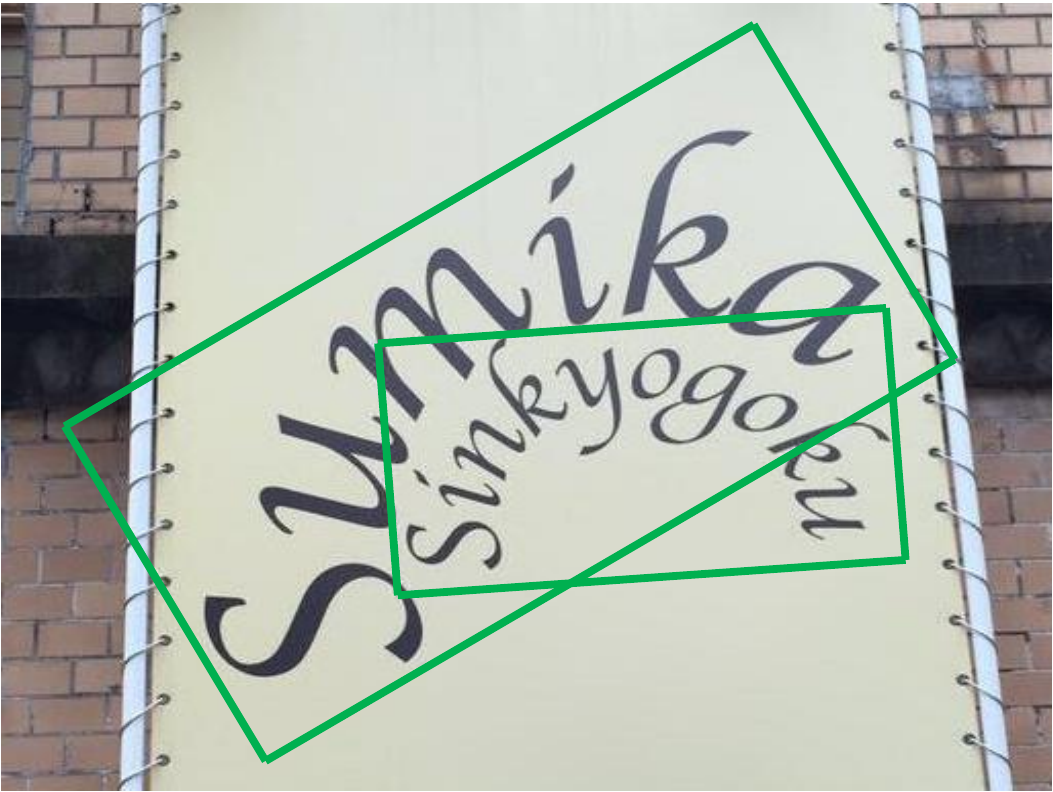}
\caption{Rotated rectangle}
\end{subfigure}
\begin{subfigure}[b]{0.32\linewidth}
\centering
\includegraphics[width=1.0\linewidth]{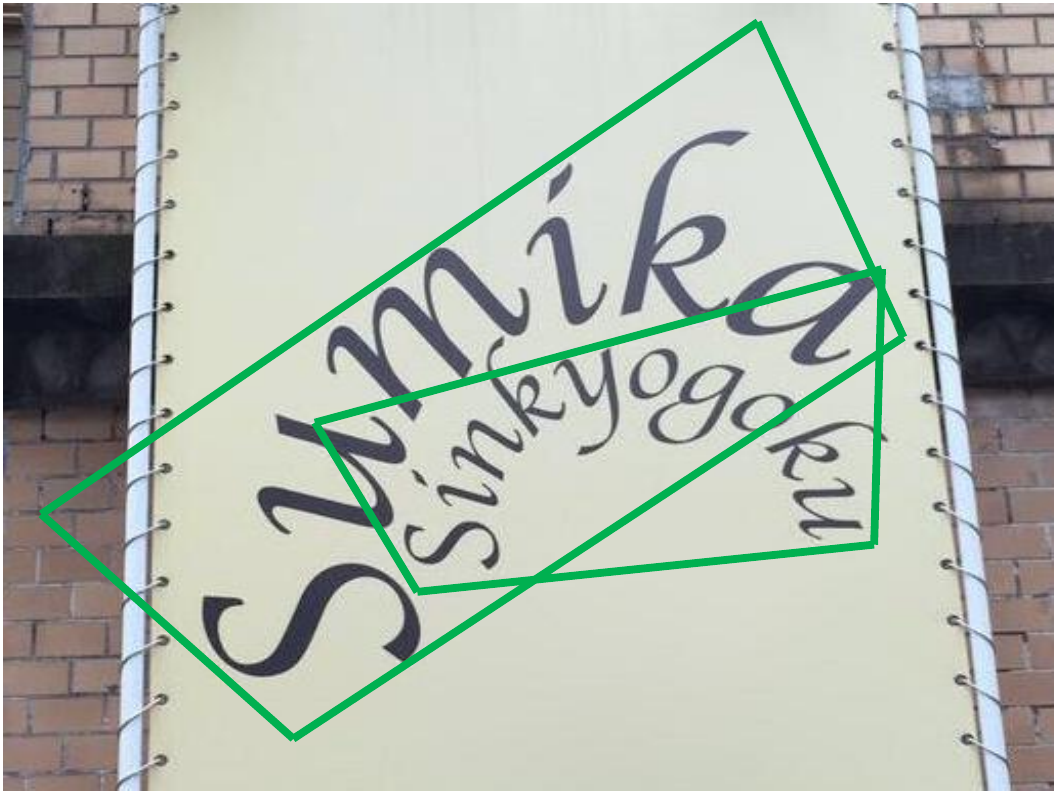}
\caption{Quadrilateral}
\end{subfigure}
\vskip 0.2cm
\begin{subfigure}[b]{0.32\linewidth}
\centering
\includegraphics[width=1.0\linewidth]{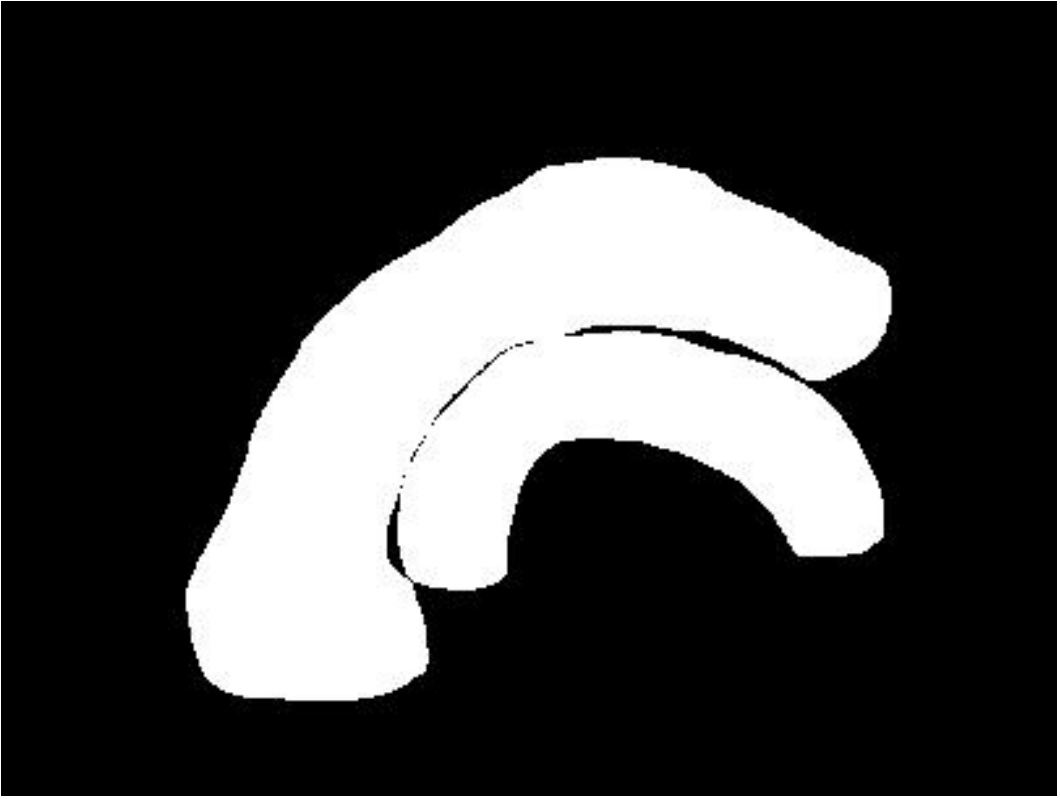}
\caption{Text mask}
\end{subfigure}
\begin{subfigure}[b]{0.32\linewidth}
\centering
\includegraphics[width=1.0\linewidth]{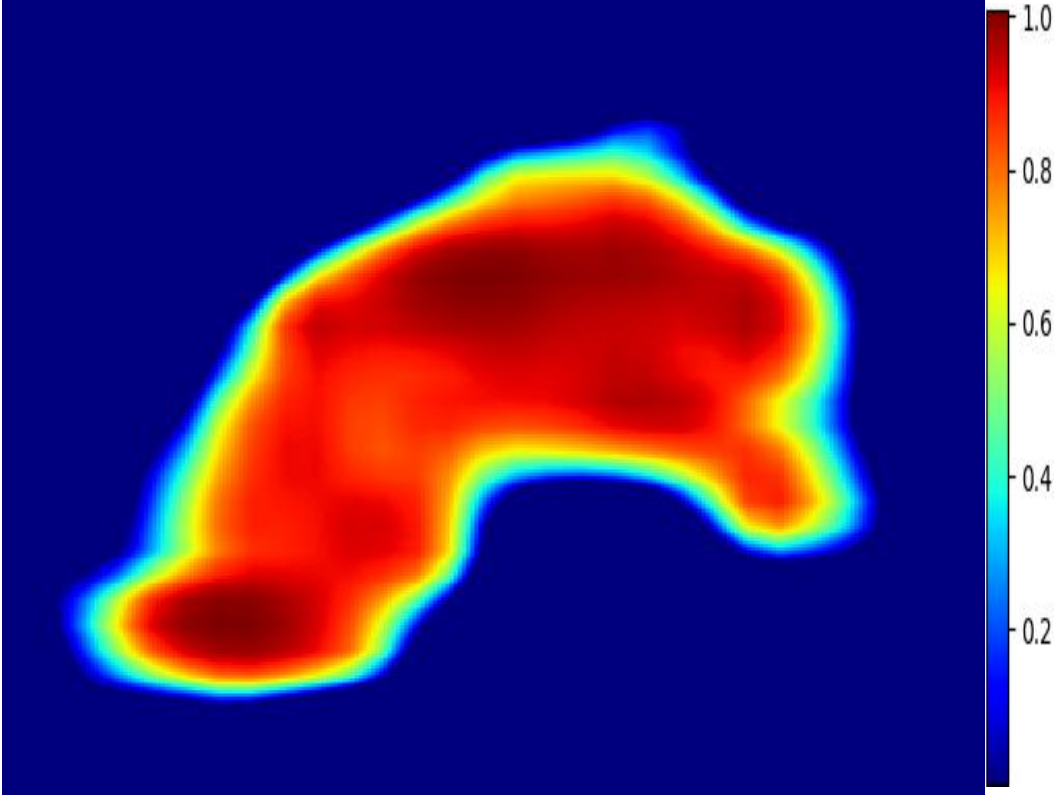}
\caption{Predicted mask}
\end{subfigure}
\begin{subfigure}[b]{0.32\linewidth}
\centering
\includegraphics[width=1.0\linewidth]{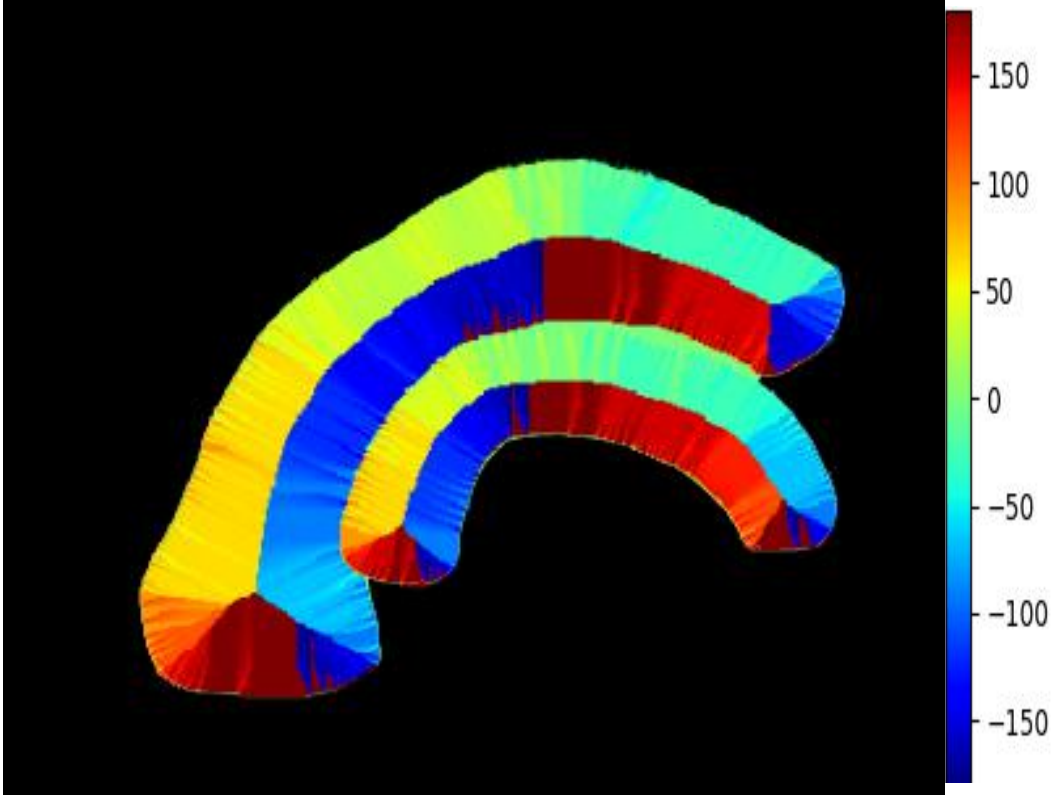}
\caption{Direction field}
\end{subfigure}

\caption{Different text representations. Classical relatively simple text representations in (a-c) fail to accurately delimit irregular texts. The text instances in (e) stick together using binary text mask representation in (d), requiring heavy post-processing to extract text instances. The proposed direction field in (f) is able to precisely describe irregular text instances.}
\label{fig:diffrep}
\end{figure}

\begin{figure*}
\centering
\includegraphics[width=0.32\linewidth]{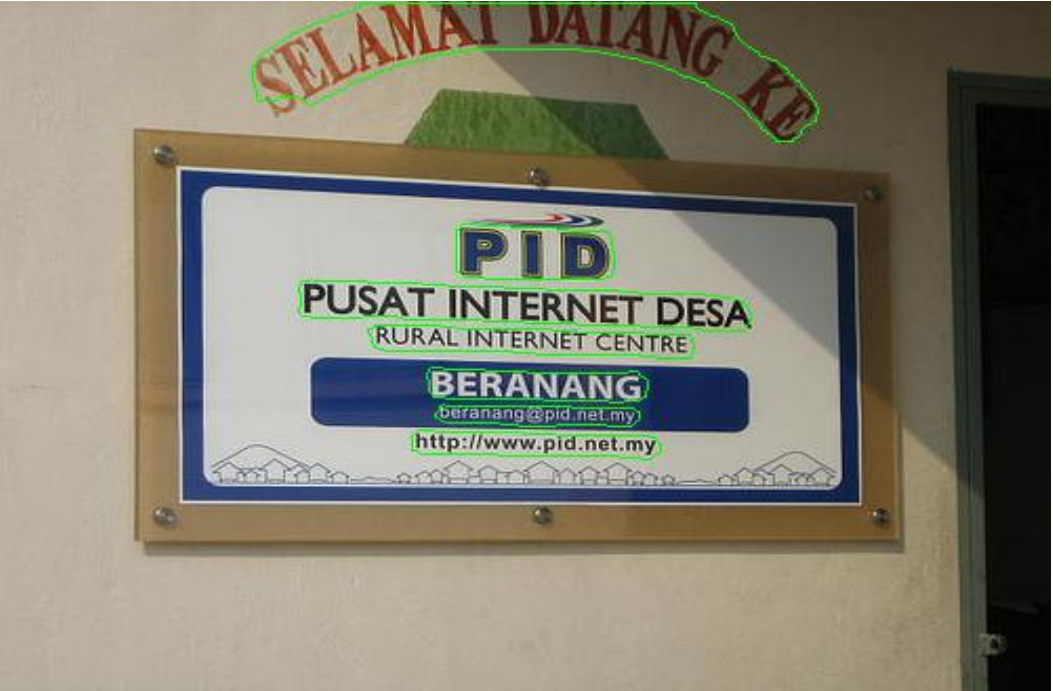}
\includegraphics[width=0.29\linewidth]{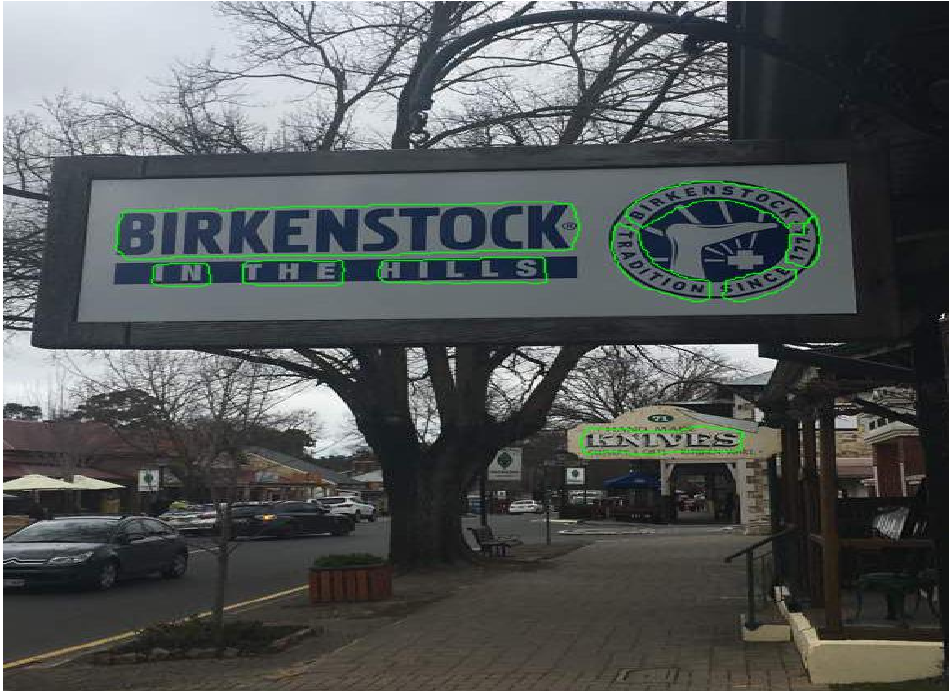}
\includegraphics[width=0.32\linewidth]{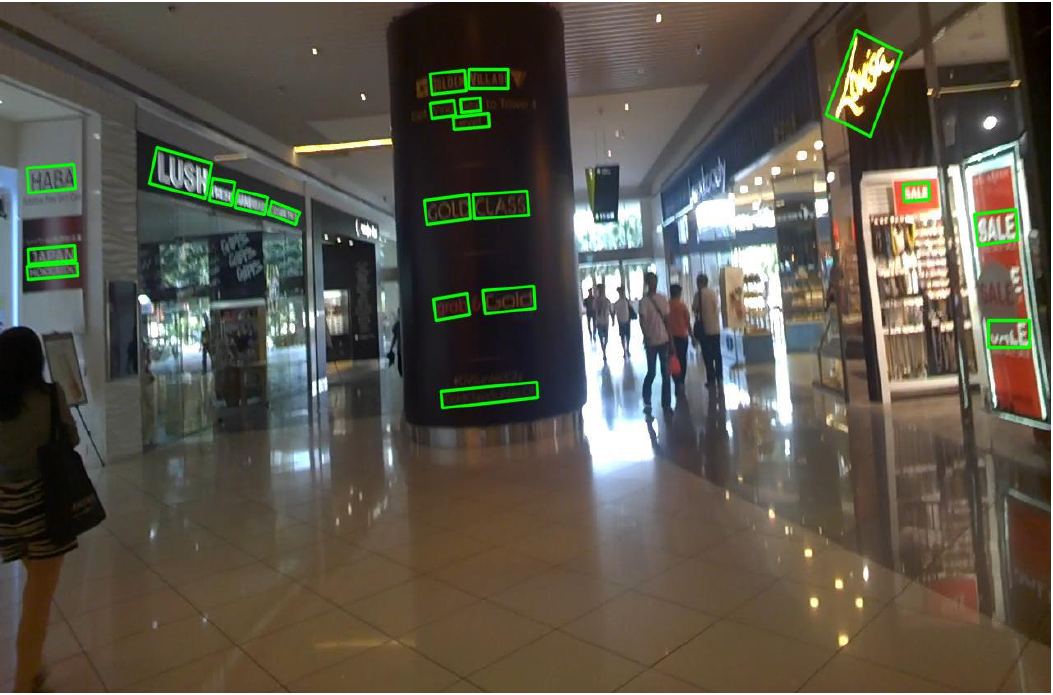}
\caption{Some irregular scene text detection results (enclosed by green contours) on some challenging images.}
\label{fig:detexamples}
\end{figure*}

The popular regression-based methods and existing hybrid  methods~\cite{tian2016ctpn,shi2017seglink,zhou2017east,liu2017deep,he2017DDR,liao2017textboxes,he2017single,hu2017wordsup,ma2018arbitrary,liao2018textboxes++,liao2018rotation,wang2018geometry,lyu2018multi,liu2018learning} achieve excellent performances on standard benchmarks. Yet, they have a strong bottleneck which assumes a text instance has a linear shape, thus adopting relatively simple text representation in terms of horizontal/oriented rectangles or quadrilaterals. Their performances drop significantly for detecting text of irregular shapes, \eg, curved text. Besides, as depicted in Fig.~\ref{fig:diffrep}(a-c), the traditional simple text representations do not achieve precise text delimitation providing texts' geometrical properties, which are useful for the subsequent recognition~\cite{shi2017end,shi2018aster}. Segmentation-based methods~\cite{zhang2016textfcn,yao2016scene,wu2017self,he2017multi,ch2017total} may not suffer from this problem. Yet, as depicted in Fig.~\ref{fig:diffrep}(e), though the predicted text region is a good estimation of text areas, it is rather difficult to separate close text instances. Indeed, many efforts of segmentation-based methods focus on how to separate segmented text regions into text instances.

In real-world scenarios, curved texts appear frequently~\cite{fabrizio2016textcatcher} and can be easily found in real life scenes such as bottles, spherical objects, clothes, logos, signboards. In two recently released datasets (Total-Text~\cite{ch2017total} and SCUT-CTW1500~\cite{liu2017detecting}) for scene text detection, around $40\%$ text instances are curved texts.

In this paper, we propose a novel text detector deemed TextField for detecting texts of arbitrary shapes and orientations. Inspired by component tree representation~\cite{salembier.1998.itip,matas2004robust,najman2006building,carlinet2014comparative} that links neighboring pixels following their intensity order to form candidate characters, we propose to learn a deep direction field, which is similar to the notion of flux image~\cite{wang2018deepflux}, to link neighboring pixels and form candidate text parts. The learned direction information is further used to group text parts into text instances. For that, the text areas are translated into text direction field first, pointing away from the nearest text boundary to each text point. Specifically, this direction field is encoded by an image of two-dimensional vectors for network training. For text areas, the field is defined as a unit vector encoding the direction, and for non-text areas, the direction field is set to $(0, 0)$. Thus, the magnitude information provides the text mask, while the direction information facilitates the post-processing of separating predicted text areas into text instances. An example of such direction field is given in Fig.~\ref{fig:diffrep}(f). We adopt a fully convolutional network to directly regress the direction field. The candidate text pixels are then obtained by thresholding the magnitude. The direction information is used to extract text instances from candidate text pixels via some morphological operators. This results in detections with precise delimitation of irregular scene texts. Several examples are given in Fig.~\ref{fig:detexamples}. The proposed TextField significantly outperforms other methods by 28\% and 8\% in F-measure on Total-Text~\cite{ch2017total} and SCUT-CTW1500~\cite{liu2017detecting}, respectively, while achieving very competitive performances on two widely adopted multi-oriented text datasets.

The main contributions of this paper are three folds: 1) We propose a novel direction field which can represent scene texts of arbitrary shapes. This direction field encodes both binary text mask and direction information facilitating the subsequent text grouping process. 2) Based on the direction field, we present a text detector named TextField, which efficiently detects irregular scene texts. 3) The proposed TextField significantly outperforms state-of-the-art methods on two curved text datasets and achieves competitive performances on two widely adopted multi-oriented text datasets.

The rest of this paper is organized as follows. We shortly review some related works on scene text detection in Section~\ref{sec:relatedworks}. The proposed method is then detailed in Section~\ref{sec:method}, followed by extensive experimental results in Section~\ref{sec:experiments}. Finally, we conclude and give some perspectives in Section~\ref{sec:conclusion}.

\section{Related works}
\label{sec:relatedworks}

Scene text detection has been extensively exploited recently. We first review some representative methods in Section~\ref{subsec:textdetection}. A comprehensive review of recent scene text detectors can be found in~\cite{ye2015text, zhu2016scene}. The comparison of the proposed TextField with some related works is depicted in Section~\ref{subsec:comparison}.

\subsection{Scene text detection}
\label{subsec:textdetection}

Scene text detection methods can be roughly classified into specifically engineered and deep learning-based methods. Before the era of deep learning, scene text detector pipelines usually consist of text component extraction and filtering, component grouping, and candidate filtering. The key step is extracting text components based on some engineered features. Maximally Stable Extremal Regions (MSER)~\cite{matas2004robust} and Stroke Width Transform (SWT)~\cite{epshtein2010detecting} are two representative works for text component extraction. Many traditional methods~\cite{wang2010word,neumann2012real,huang2013text,huang2014robust, yin2015multi} are based on these two algorithms. Other examples of this type are~\cite{gomez2013multi,li2014characterness}. Most recent methods shift to deep neural networks to extract scene texts. In general, they can be roughly summarized into regression-based, segmentation-based, and hybrid methods. For the regression-based ones, they can be further divided into two categories based on the target to regress: proposal-based and part-based methods.

\textbf{Proposal-based methods:} Proposal-based methods are mainly inspired by recent object detection pipelines. TextBoxes~\cite{liao2017textboxes} directly adapts SSD~\cite{liu2016ssd} for scene text detection by using long default boxes and convlutioinal filters to cope with the significantly varied aspect ratios. TextBoxes++~\cite{liao2018textboxes++} extends TextBoxes by regressing quadrilaterals instead of horizontal bounding boxes. Ma {\em et al.}~\cite{ma2018arbitrary} attempt to solve the multi-oriented text detection by adopting Rotated Regional Proposal Network (RRPN) in the pipeline of faster r-cnn. Quadrilateral sliding windows are adopted in~\cite{liu2017deep} to detect multi-oriented texts. Wordsup~\cite{hu2017wordsup} explores word annotations for character-based text detection. SSTD~\cite{he2017single} introduces the attention mechanism by FCN to suppress background interference, improving accurate detection of small texts. In~\cite{liao2018rotation}, Liao {\em et al.} propose to apply rotation-invariant and sensitive features for text/non-text box classification and regression, respectively, boosting long multi-oriented text detection. Wang et al.~\cite{wang2018geometry} propose instance transformation network by considering geometry-aware information for scene text detection.


\textbf{Part-based methods:} Some other regression-based methods tend to regress text parts while predicting the linking relationship between them. In~\cite{tian2016ctpn}, the authors propose a Connectionist Text Proposal Network (CTPN) by first predicting vertical text parts, then adopting a recurrent neural network to link text parts. Shi {\em et al.} present a network named SegLink~\cite{shi2017seglink} to first detect text parts named text segments while predicting the linking relationship between neighboring text segments. A novel framework named Markov Clustering Network (MCN) is proposed in~\cite{liu2018learning}. In this work, the authors propose to regard an image as a stochastic flow graph, where the flows are strong between text nodes (\ie, text pixels) but weak for the others. Then a Markov clustering process is applied to form text instances from the predicted flow graph. In~\cite{lyu2018multi}, {\em Lyu et al.} propose to first regress four corners of text boxes, followed by a combination of corners and Non-Maximum Suppression (NMS) process to achieve accurate multi-oriented text localization.

\textbf{Segmentation-based methods:} Segmentation-based approaches regard text detection as a text area segmentation problem, which is usually achieved via Fully Convolutional Neural Network (FCN). They mainly differ in how to post-process the predicted text regions into words or text lines. In~\cite{zhang2016textfcn}, Zhang {\em et al.} adopted an FCN to estimate text blocks, on which candidate characters are extracted using MSER. Then they use traditional grouping and filtering strategies to achieve multi-oriented text detection. In addition to text block (word or line) prediction, Yao {\em et al.}~\cite{yao2016scene} also propose to predict both individual characters and the orientation of text boxes via an FCN in a holistic fashion. Then a grouping process based on the three estimated properties of text yields the text detection. Ch'ng {\em et al.}~\cite{ch2017total} fine-tune DeconvNet~\cite{noh2015learning} to achieve curved text detection. In~\cite{he2017multi}, the authors consider the text detection as an instance segmentation problem using multi-scale image inputs. They adopt an FCN to predict text blocks, followed by two CNN branches predicting text lines and instance-aware segmentations from the estimated text blocks. Wu {\em et al.}~\cite{wu2017self} introduce text border in addition to text/non-text segmentation, which results in a three-class semantic segmentation, facilitating the separation of adjacent text instances. Xue {\em et al.} further improve~\cite{wu2017self} by exploiting bootstrapping techniques and designing semantics-aware text border detection technique for accurate text localization.

\textbf{Hybrid methods:} It is also worth to mention that some other methods leverage segmentation to classify text/non-text pixels and then localize texts via bounding box regression. For example, East~\cite{zhou2017east} and Deep regression~\cite{he2017DDR} both perform per-pixel rotated rectangle or quadrilateral estimation.

\begin{figure*}[!ht]
\centering
\includegraphics[width=0.96\linewidth]{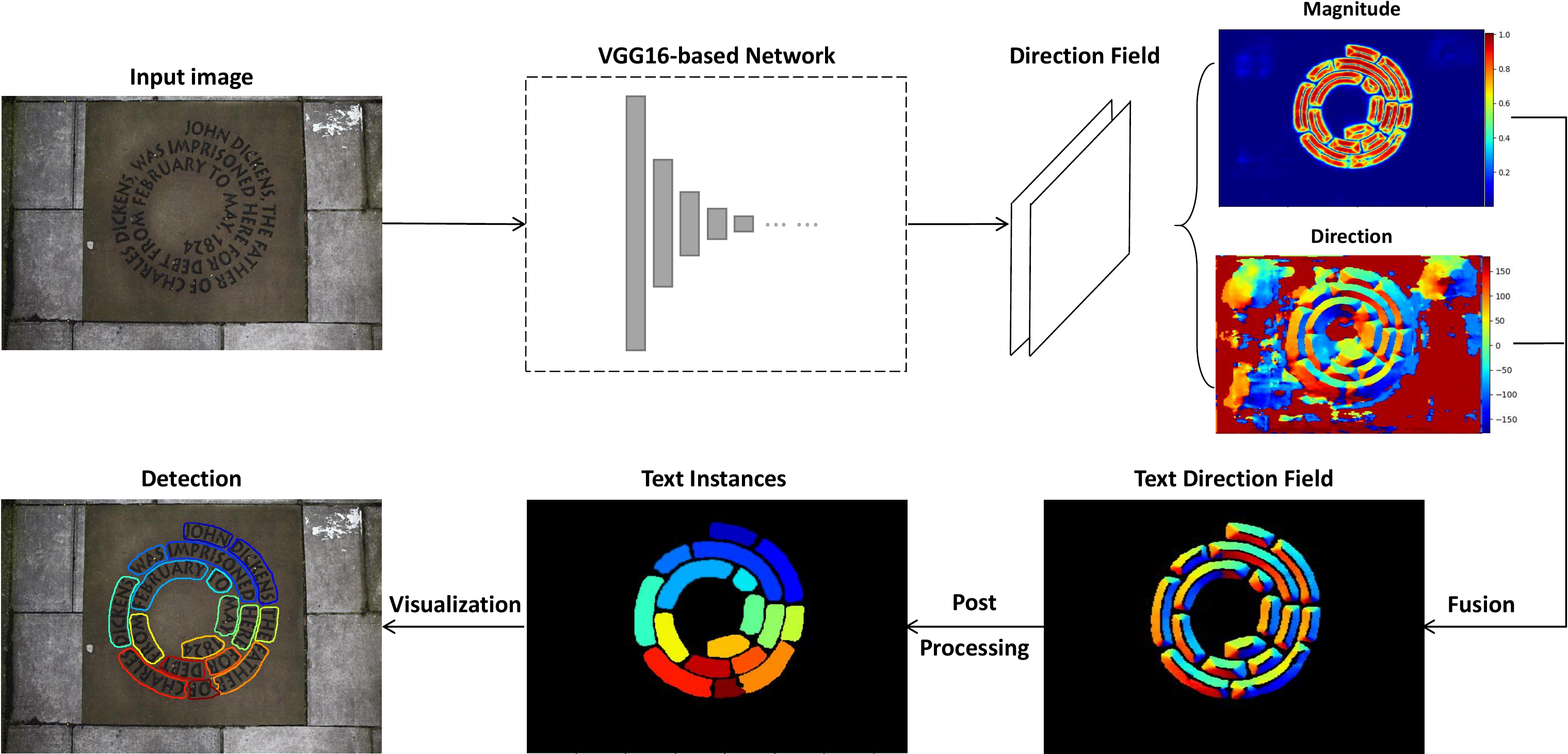}
\vskip 0.4cm
\caption{Pipeline of the proposed method. Given a test image, the network predicts a novel direction field in terms of a two-channel map, which can be regarded as an image of two-dimensional vectors. To better show the predicted direction field, we calculate and visualize its magnitude and direction information. Text instances are then obtained based on these information via the proposed post-processing using some morphological tools.}
\label{fig:pipeline}
\end{figure*}

\subsection{Comparison with related works}
\label{subsec:comparison}

\textbf{TextField Versus Traditional component-based methods:} Traditional methods rely on engineered features to extract text components, and heuristic grouping rules to form text instances. Each module requires careful parameter tuning, resulting in sub-optimal performance and slow runtime of the whole pipeline. The proposed TextField leverages deeply learned direction field which encodes both text mask and direction information facilitating subsequent text grouping process. The whole pipeline is more effective in both performance and runtime.

\textbf{TextField Versus Proposal-based and hybrid} methods: The proposal-based and hybrid scene text detection methods are mainly inspired by recent object detection pipelines, which have relatively less flexible text representations. They usually regress text instances in form of horizontal/oriented rectangles or quadrilaterals, having limited ability in detecting irregular texts (\eg, curved texts). TextField does not suffer from this limitation. Benefiting from the proposed direction field, TextField is able to accurately detect texts of irregular shapes.

\textbf{TextField Versus Part-based methods:} Part-based methods decompose the text instances into text parts, then attempt to link the neighboring text parts. They enjoy a more flexible representation, and can somehow alleviate the problem of relatively simple text representation inherited in proposal-based methods. Yet, driven by the employed linking or combination strategy, these methods usually produce multi-oriented text detections. The proposed direction field is versatile in representing multi-oriented and curved texts, making TextField perform equally well in detecting any irregular texts.

\textbf{TextField Versus Segmentation-based methods:} Due to the significantly varied sizes and aspect ratios, most segmentation-based methods are built upon semantic segmentation, followed by a heavy post-processing step to separate the predicted text areas into text instances. In addition to text mask, some information such as text border, text line, text box orientation, or linking relationship between neighboring pixels is also predicted to ease the separation of adjacent texts. Yet, such additional information either limits the method to multi-oriented text detection or also faces similar problem with text semantic segmentation in separating adjacent texts. TextField directly regresses the direction field which encodes both text mask and direction information that points away from text boundary, thus allowing simple separation of adjacent texts. In this sense, TextField is more elegant and efficient in detecting irregular texts.

It is worth to mention that direction information has also been diversely exploited in some other applications~\cite{bai2017directional,bai2017deep,chen2018masklab}, which involve different definitions or usages.

\section{Proposed Methodology}\label{sec:method}

\begin{figure*}[!ht]
\centering
\includegraphics[width=0.96\linewidth]{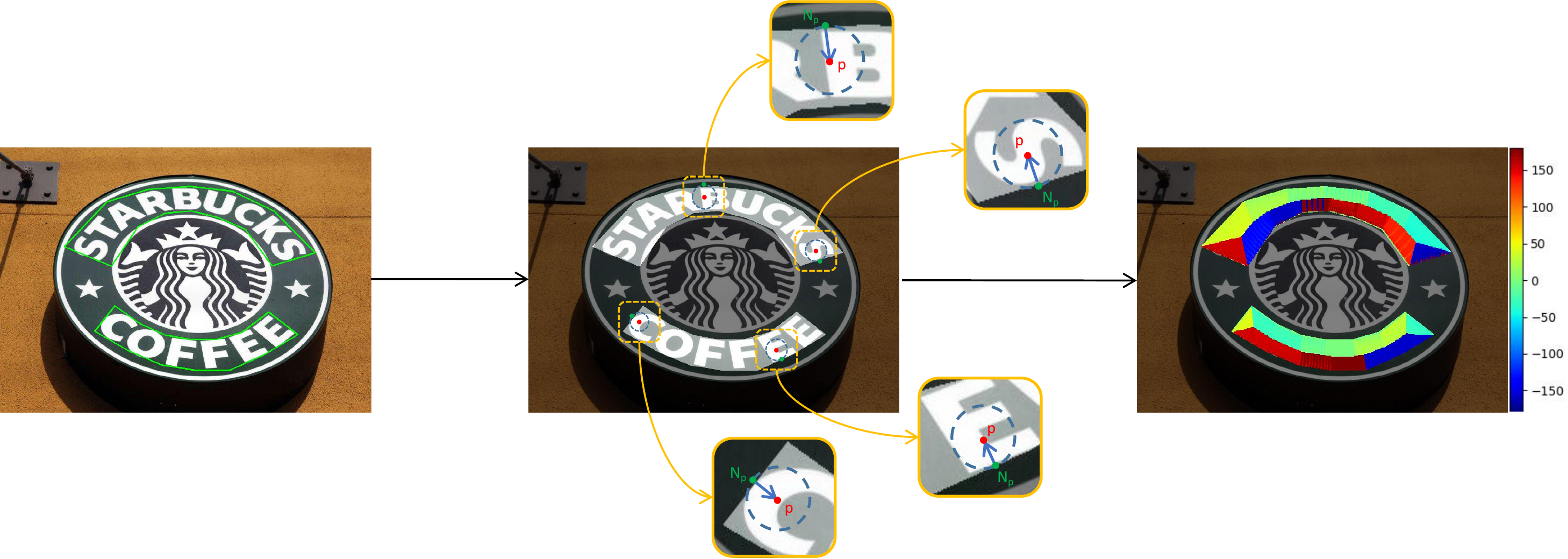}
\vskip 0.2cm
\caption{Illustration of the proposed direction field. Given a training image and its text annotation, a binary text mask can be easily generated. For each text pixel $p$, we find its nearest non-text pixel $N_p$. Then, a two-dimensional unit vector that points away from $N_p$ to $p$ is defined as the direction field on $p$. For non-text pixels, the direction field is set to $(0, 0)$. On the right, we visualize the direction information of the text direction field.}
\label{fig:directionfield}
\end{figure*}

\subsection{Overview}
\label{subsec:overview}


The proposed method relies on a fully convolutional neural network to produce a dense per-pixel direction field for detecting irregular texts. The pipeline is depicted in Fig.~\ref{fig:pipeline}. In general, we regard the text detection problem as text instance segmentation. For that, we propose a novel direction field, aiming at segmenting texts and also separating adjacent text instances. More specifically, for a text pixel $p$, its direction field is represented by a two-dimensional unit vector that points away from its nearest text boundary pixel. This direction field is detailed in Section ~\ref{subsec:directionfield}. Benefiting from such novel representation, the proposed TextField can easily separate text instances that lie close to each other. Furthermore, such direction field is appropriate for describing text of arbitrary shapes. We adopt a VGG16-based network to learn the direction field. To preserve spatial resolution and take full advantage of multi-level information, we exploit a widely used multi-level feature fusion strategy. The network architecture is presented in Section~\ref{subsec:networkarchitecture}. Some specific adaptions for the network training are given in Section~\ref{subsec:optimization}, including online hard negative mining and a weighted loss function for our per-pixel regression task. Both adaptions are dedicated to force our network to focus more on hard pixels and eliminate the effects caused by quantitative imbalance between foreground and background pixels. Finally, a novel post-processing based on mathematical tools (see Section~\ref{subsec:postprocessing}) is proposed to group pixels, forming detected text instances thanks to the predicted text direction field.

\begin{figure}
\centering
\includegraphics[width=0.93\linewidth]{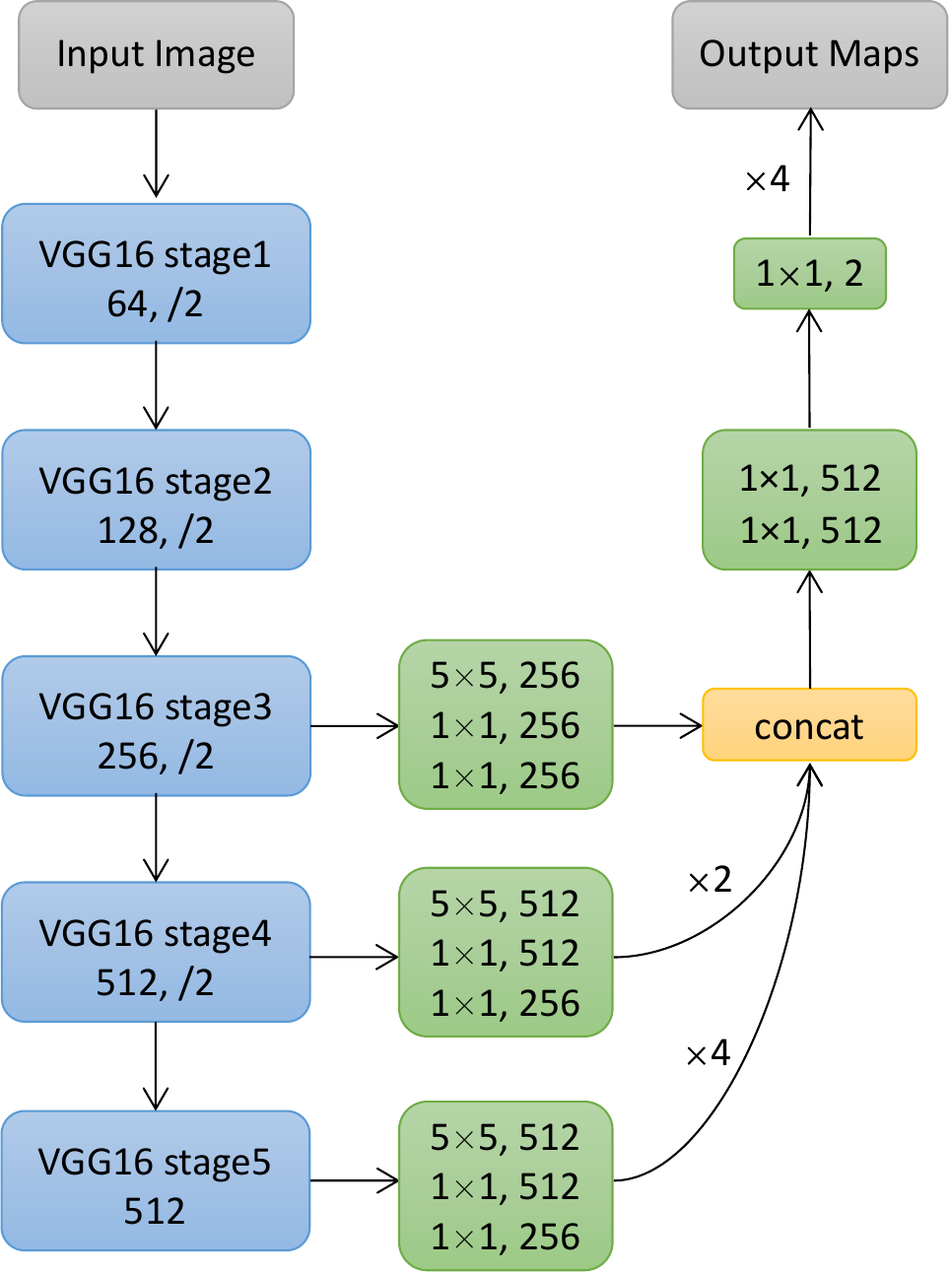}
\vskip 0.2cm
\caption{Network architecture. We adopt the pre-trained VGG16~\cite{vgg16network} as the backbone network and multi-level feature fusion to capture multi-scale text instances. The network is trained to predict dense per-pixel direction field.}
\label{fig:networkarchitecture}
\end{figure}

\subsection{Direction field}
\label{subsec:directionfield}

As pointed out in Sec.~\ref{subsec:comparison}, though proposal-based and part-based text detectors have achieved impressive performances on multi-oriented text detection, they do not perform well for curved texts. Segmentation-based approaches can somehow tackle this limitation via binary text mask (of arbitrary shapes) segmentation. Yet they can hardly separate adjacent text instances. To address these issues, we propose a novel direction field for detecting irregular texts.

Instead of binary text mask involved in the segmentation-based approaches, we propose the direction field that encodes both binary text mask and direction information that can be used to separate adjacent text instances. As illustrated in Fig.~\ref{fig:directionfield}, for each pixel $p$ inside a text instance $T$, let $N_p$ be the nearest pixel to $p$ lying outside the text instance $T$, we then define a two-dimensional unit vector $V_{gt}(p)$ that points away from $N_p$ to the underlying text pixel $p$. This unit vector $V_{gt}(p)$ directly encodes approximately relative location of $p$ inside $T$ and highlights the boundary between adjacent text instances. For the non-text area, we represent those pixels with $(0,0)$. Formally, the proposed direction field is given by:
\begin{equation}
	V_{gt}(p) \; = \; \revise{\left\{\begin{matrix}
\ \overrightarrow{N_p p}/\left\vert\overrightarrow{N_p p}\right\vert, & p\in \mathbb{T}  \\ \\
(0,0), & p \not\in \mathbb{T}
\end{matrix}\right.}
\label{eq:textrep}
\end{equation}
where $\left\vert\overrightarrow{N_p p}\right\vert$ denotes length of the vector starting from pixel $N_p$ to $p$, and $\mathbb{T}$ stands for all the text instances in an image. In practice, for each text pixel $p$, it is simple to compute its nearest pixel $N_p$ outside the text instance containing $p$ by distance transform algorithm. Consequently, it is rather straightforward to transform a traditional text annotation to the proposed direction field.

The proposed direction field given by Eq.~\eqref{eq:textrep} is appropriate for detecting irregular texts. In fact, the magnitude of direction field $V$ is equivalent to binary text mask. Thus, we rely on magnitude of $V$ to differentiate text and non-text pixels. The direction information encoded in $V$ facilitates the separation of adjacent text instances (see Sec.~\ref{subsec:postprocessing}).



\begin{figure*}
\centering
\includegraphics[width=0.96\linewidth]{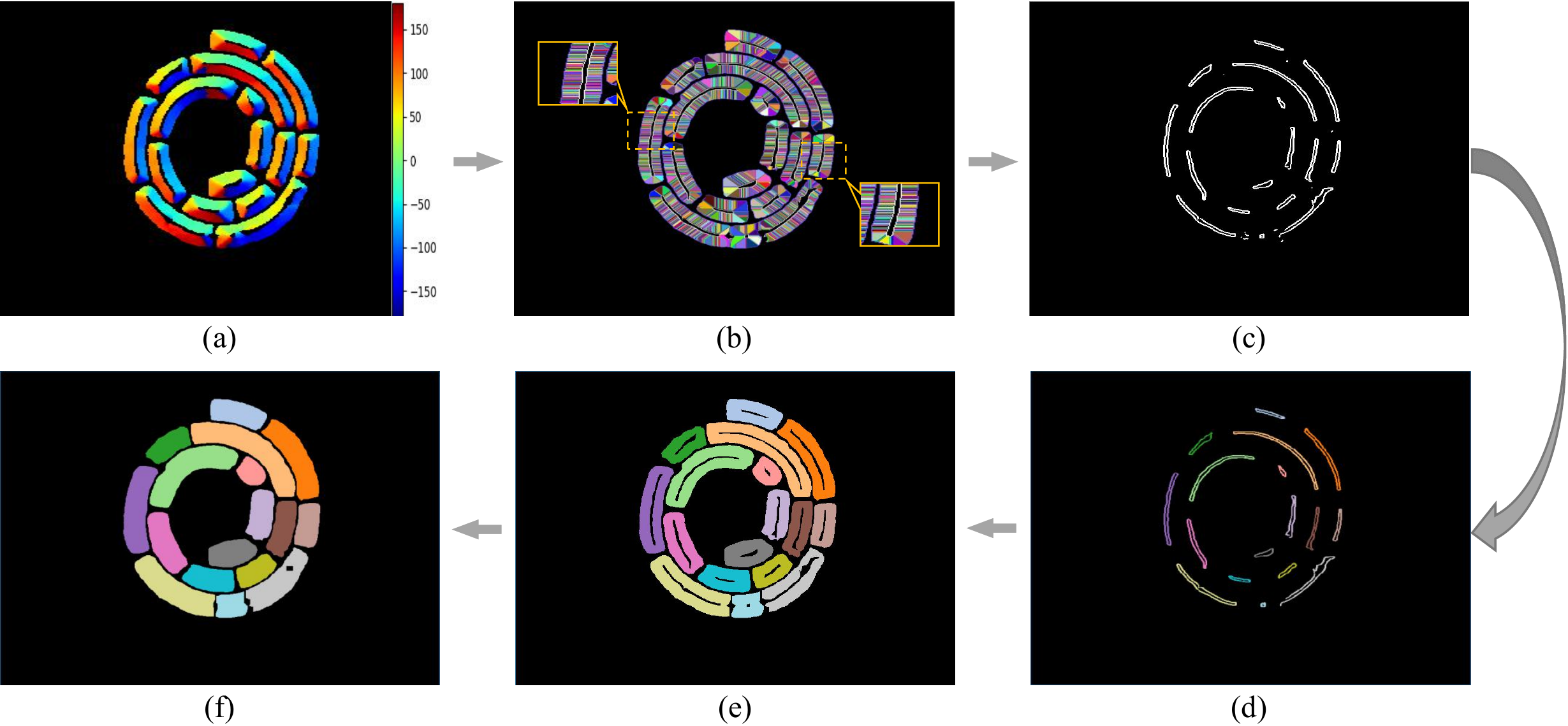}
\caption{Illustration of the proposed post-processing on a test image. (a): Directions on candidate text pixels; (b): Text superpixels (in different color) and their representatives (in white); (c): Dilated and grouped representatives of text superpixels; (d): Labels of filtered representatives; (e): Candidate text instances; (f) Final segmented text instances.}
\label{fig:postprocessing}
\end{figure*}

\subsection{Network architecture}
\label{subsec:networkarchitecture}

The proposed network architecture to learn the direction field for detecting irregular texts is depicted in Fig.~\ref{fig:networkarchitecture}. We adopt a fully convolutional neural network which mainly consists of two parts: feature extraction and multi-level feature fusion. The backbone network to extract features is the VGG16 network~\cite{vgg16network} pre-trained on ImageNet~\cite{imagenet}. We discard the last pooling layer and its following fully connected layers. Since text sizes may vary significantly, it is difficult to detect small text instances with only coarse features. Therefore, we merge features from different stages to capture multi-scale text instances. More specifically, we exploit the feature maps from $stage3$, $stage4$, and $stage5$ of the VGG16 backbone network. These multi-level features are upsampled to the same size as the feature map from $stage3$, and are then merged together by concatenation. This is followed by three convolution layers, resulting in a two-channel map that predicts the direction field given by Eq.~\eqref{eq:textrep}. Finally, we upsample the predicted direction field to the original size. We adopt bilinear interpolation for all the upsampling operations.

It is worth to note that the proposed method is not severely bottlenecked by the limited receptive field. In fact, the proposed direction field in Eq.~\eqref{eq:textrep} only relies on local clues (\ie, location of the nearest text boundary). Thus, we only require a receptive field that covers the short side of text instance. Whereas, for the classical proposal-based methods, a receptive field larger than the long side of underlying text instance is usually needed. Consequently, the proposed method is more flexible in detecting irregular long texts.

\subsection{Optimization}
\label{subsec:optimization}

\subsubsection{Training objective}
\label{subsubsec:trainingobjective}

We leverage the network depicted in Section~\ref{subsec:networkarchitecture} to regress the proposed direction field. The network parameters are optimized with an instance-balanced Euclidean loss. More specifically, the loss function to minimize is a weighted sum of the mean squared error on each pixel of the image domain $\Omega$. This is given by:
\begin{equation}
L \; = \; \sum_{p \in \Omega}{w(p) * \left\|V_{gt}(p)-V_{pred}(p)\right\|_2},
\label{eq:loss}
\end{equation}
where $V_{pred}$ is the predicted direction field, and $w(p)$ denotes the weight coefficient of pixel $p$. Since text sizes may vary significantly in scene images, if all text pixels contribute equally to the loss function, large text instances will be dominative in the loss computation while small ones will be ignored. To tackle this problem, we adopt an instance-balanced strategy. More precisely, for an image containing $N$ text instances, the weight $w$ for a given pixel $p$ is defined as follows:
\begin{equation}
\revise{w(p)=}\revise{\left\{\begin{matrix}
\ \frac{\sum_{T \in \mathbb{T}} |T|}{N*|T_p|}, & p \in \mathbb{T}  \\ \\
1, & p \not\in \mathbb{T}
\end{matrix}\right.}
\label{eq:weight}
\end{equation}
where $|T|$ denotes the total number of pixels in a text instance $T$, and $T_p$ stands for the text instance containing pixel $p$. In this way, each text instance of any size is endowed with the same weight, contributing equally to the loss function in Eq.~\eqref{eq:loss}. This is consistent with current text detection system such that each text instance is equally important.

\vskip 1mm
\subsubsection{Online hard negative mining}
\label{subsubsec:OHEM}

In scene images, text instances usually occupy a small area of the image. Thus, the number of text pixels and non-text pixels is rather imbalanced. To alleviate this problem and to make the network training focus more on pixels which are hard to distinguish, we adopt hard mining following the online hard negative mining strategy proposed in~\cite{shrivastava2016OHEM}. More specifically, non-text pixels are sorted in a decreasing order of their per-pixel loss. Then only the front $\gamma* (\sum_{T \in \mathbb{T}} |T|)$ non-text pixels are reserved for backpropagation, where $\gamma$ is a given hype-parameter that denotes the ratio of non-text pixels with respect to the total number of text pixels when computing the total loss.

\subsection{Inference and post-processing}
\label{subsec:postprocessing}

For a given image, the trained network predicts the direction field in terms of 2D vectors. We propose a novel post-processing pipeline using some morphological tools to obtain the final text detection results from this prediction. Precisely, as described in Section~\ref{subsec:directionfield}, the magnitude of the predicted direction field $V_{pred}$ highlights text/non-text areas. Thus, we first threshold the magnitude image with a thresholding value $\lambda_m$ to obtain candidate text pixels $C$. It is worth to note that pixels lying around text symmetrical axis usually have low magnitude due to the cancellation of opposite direction in learning and upsampling. The text detection problem then amounts to group candidate text pixels into text instances. For that, we first segment the candidate text areas into text superpixels (\ie, text parts depicted in different color in Fig.~\ref{fig:postprocessing}(b)), which are then grouped together to form candidate text instances. A last text instance filtering step is adopted to yield the final detected texts. This process is depicted in Fig.~\ref{fig:postprocessing} and Algorithm~\ref{algo:textseg}, and summarized in the following:

\textbf{Text superpixel segmentation:} The magnitude information of the predicted direction field $V_{pred}$ is used to generate candidate text pixels. Then we rely on the direction information carried by $V_{pred}$ to segment the candidate text areas into text superpixels. Precisely, for each candidate text pixel $p$, the direction information $\angle V_{pred}(p)$ is binned into one of the 8 directions, pointing to its nearest neighboring candidate text pixel denoted as $\mathcal{P}(p)$, standing for the parent of pixel $p$. Each candidate text pixel points to a unique neighboring pixel. Consequently, the parent image $\mathcal{P}$ forms a forest structure $\mathcal{F}$, partitioning the candidate text areas into text superpixels, each of which is represented by a tree $\mathcal{T} \in \mathcal{F}$. This text superpixel segmentation can be efficiently achieved by blob labeling algorithm (see line 7-15 in Algorithm~\ref{algo:textseg}).

\textbf{Text superpixel grouping:} Based on the segmented text superpixels represented by trees, we propose a simple grouping method to form candidate text instances. Since the proposed direction field encodes the direction away from the nearest boundary, the root pixels of all trees locate near the symmetry axis of each text instance.  We consider all these root pixels as the representatives of all the text superpixels. The representatives of a text instance usually are close to each other (See Fig.~\ref{fig:postprocessing}). We apply a simple dilation $\delta$ (with $3 \times 3$ structuring element) to group the representatives of the same text instance. This is followed by a connected component labeling that forms candidate text instances. The text superpixel grouping is depicted in line 17-21 of Algorithm~\ref{algo:textseg}.

\textbf{Text instance filtering:}
After the extraction of candidate text instances, we apply some filtering strategies to get rid of some non-text instances following their shapes and sizes. As illustrated in Fig.~\ref{fig:postprocessing}, the representative pixels of a text instance should have a symmetrical distribution of directions. Therefore, all the representative pixels of a text instance should be approximately paired in the sense of having opposite directions. Based on this observation, we count the ratio of non-paired representatives, and filter out the candidate text instances having a ratio lower than a given value $\lambda_r$ (set to 0.6). For the remaining candidate text instances, we apply a morphological closing $\phi$ (with $11 \times 11$ structuring element) to fill the inside holes. Then we also discard some noisy candidate instances whose areas are smaller than $\lambda_a$ (set to 200). The remaining candidate text instances are the final detected texts. The text instance filtering is given in line 23-27 in Algorithm~\ref{algo:textseg}.

Specifically, the proposed post-processing is detailed in Algorithm~\ref{algo:textseg}. The core body of the algorithm is the blob labeling to construct text superpixels via the forest structure encoded by the parenthood image $\mathcal{P}$. This blob labeling process can be efficiently implemented using a stack data structure $S$ and an auxiliary image $visited$. The text superpixels are labeled by the image $\mathcal{L}$. Then we identify the representative pixels $R$ by root pixels of those trees. These representative pixels are also stored by an image $\mathcal{M}$. We then apply a dilation $\delta$ with kernel $k_1 \times k_1$ ($k_1 = 3$) to group representative pixels, followed by a connected labeling {\em CC\_Labeling} to form candidate text instances. We then filter out some candidate text instances by the ratio of non-paired representatives {\em Filter\_Unbalanced\_Text}. The label of each remaining candidate text instance is then propagated to all the pixels inside the same text superpixels. Finally, we apply a closing $\phi$ with kernel $k_2 \times k_2$ ($k_2$ = 11) to fill the holes inside each candidate text instance, followed by a removal of small candidate text instances. This post-processing gives the final detected texts encoded by $\mathcal{M}$.

\begin{algorithm}
\caption{Text inference with a morphological post-processing on predicted direction field $V_{pred}$. $\mathcal{M}$ is the final text instance segmentation map. See the corresponding texts in Section~\ref{subsec:postprocessing} for details.}
\label{algo:textseg}
\LinesNumbered
\SetAlgoLined
\SetKw{kAnd}{and}
\SetKw{kOr}{or}


{\em Text\_Inference}($V_{pred}$, $\lambda_m$, $\lambda_r$, $\lambda_a$) \\
$\mathcal{M}, \mathcal{L} \gets 0$, $l \gets 0$, $C, R, S \gets \emptyset$,  $visited \gets  \mathbf{False}$, $\mathcal{P} \gets p_0$ //initialization \;

//{\it get candidate text pixels} \\
\ForEach{$p \in \Omega$}
{
	\lIf{$|V_{pred}(p)| \geq \lambda_m$}
    {
		$C \gets C \cup p$
	}
}
//{\it blob lableing to construct trees encoded by $\mathcal{P}$} \\
\ForEach{$p \in C$ \kAnd $\mathbf{not} \; visited(p)$}
{
	$S$.push($p$), $l \gets l + 1$ \;
    \While{$S \neq \emptyset$}
    {
    	$p' \gets S.$pop(), $visited(p') \gets \mathbf{True}$,
        $\mathcal{L}(p') \gets l$ \;
        $\mathcal{P}(p') \gets \mathcal{N}_{\angle V_{pred}(p')}(p')$ \;
        \ForEach{$q \in \mathcal{N}(p')$}
        {

        	\If{$q \in C$ \kAnd $\mathbf{not} \; visited(q)$}
            {
            	\If{$q = \mathcal{N}_{\angle V_{pred}(p')}(p')$ \kOr $p' = \mathcal{N}_{\angle V_{pred}(q)}(q)$}
                {
            		$S$.push($q$) \;
                }
            }
        }
    }
}

//{\it grouping text superpixels via their representatives} \\
\ForEach{$p \in C$}
{
	\If{$\mathcal{P}(p) = p_0$}
    {
    	$R \gets R \cup p$, $\mathcal{M}(p) \gets 1$ \;
    }
}
$\mathcal{M} \gets \delta_{k_1}(\mathcal{M})$ \;
$\mathcal{M} \gets$ {\em CC\_Labeling($\mathcal{M}$)} \;

//{\it text instance filtering by the shape and size} \\
$\mathcal{M} \gets$ {\em Filter\_unblanced\_Text($\mathcal{M}$, $R$, $\lambda_r$)} \;
\ForEach{$r \in R$}
{
	$\mathcal{M} \gets$ {\em Propagate\_Label}($\mathcal{M}$, $\mathcal{L}$, $r$) \;
}
$\mathcal{M} \gets \phi_{k_2}(\mathcal{M})$ \;
$\mathcal{M} \gets$ {\em Filter\_Small\_Regions}($\mathcal{M}$, $\lambda_a$) \;

\Return{$\mathcal{M}$} \;
\end{algorithm}

\begin{figure*}
\centering
\begin{subfigure}[b]{1.0\textwidth}
\centering
\includegraphics[width=1.0\linewidth]{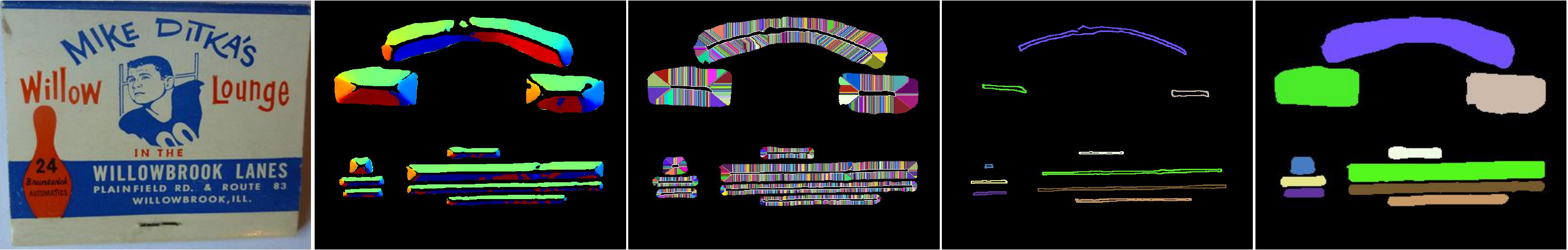}
\caption{}
\label{fig:ctwppeg}
\end{subfigure}

\vspace{1mm}
\begin{subfigure}[b]{1.0\textwidth}
\centering
\includegraphics[width=1.0\linewidth]{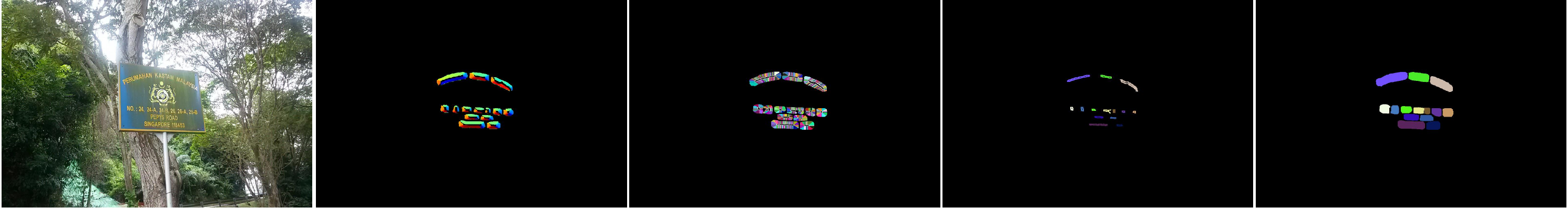}
\caption{}
\label{fig:totaltextppeg}
\end{subfigure}

\vspace{1mm}
\begin{subfigure}[b]{1.0\textwidth}
\centering
\includegraphics[width=1.0\linewidth]{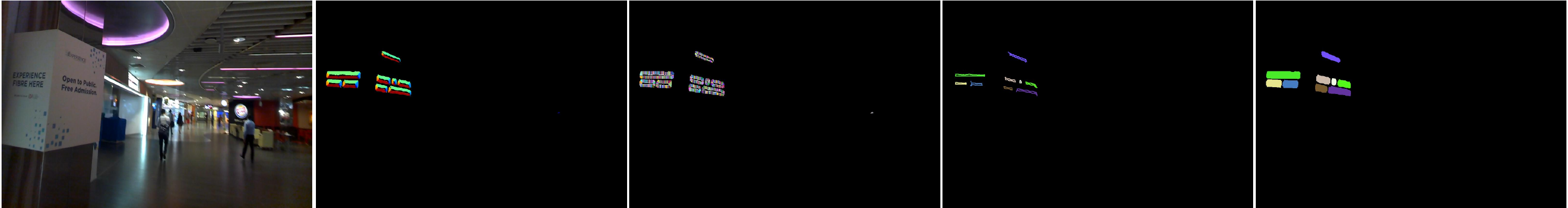}
\caption{}
\label{fig:ic15ppeg}
\end{subfigure}

\vspace{1mm}
\begin{subfigure}[b]{1.0\textwidth}
\centering
\includegraphics[width=1.0\linewidth]{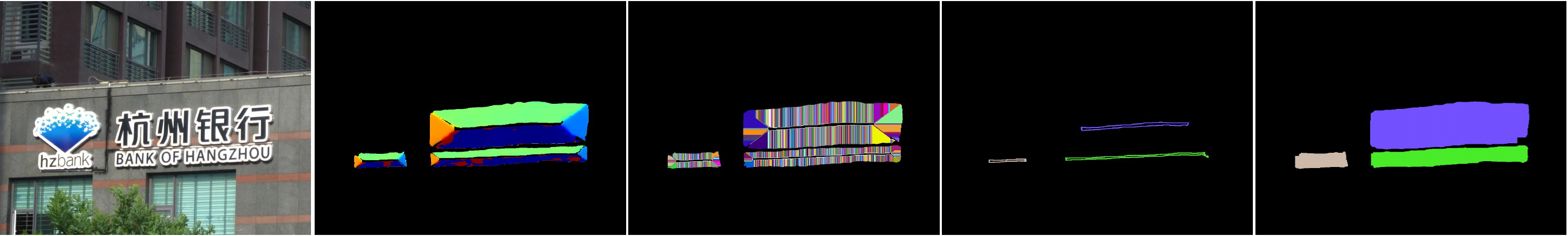}
\caption{}
\label{fig:td500ppeg}
\end{subfigure}
\caption{Visualization of learned direction field and some involved post-processing steps on test images from SCUT-CTW500 in (a), Total-Text in (b), IC15 in (c), and MSRA-TD500 in (d), respectively. From left to right: input images, directions on candidate text pixels, text superpixels (in different color) and their representatives (in white), labels (in different color) of filtered representatives, and final segmented instances.}
\label{fig:postprocessing_eg}
\end{figure*}

\section{Experiments}\label{sec:experiments}

The proposed method is appropriate for detecting irregular texts. In the following, we evaluate the proposed method on four public benchmark datasets: SCUT-CTW1500~\cite{liu2017detecting} and Total-Text~\cite{ch2017total} which contain curved texts, ICDAR2015 Incidental Scene Text (IC15)~\cite{karatzas2015icdar} and MSRA-TD500~\cite{yao2012detecting} which mainly consist of multi-oriented texts in terms of oriented rectangles or general quadrilaterals. SynthText in the Wild~\cite{gupta2016synthetic} is also adopted to pre-train the network. A short description of these datasets and adopted evaluation protocol is given in Section~\ref{subsec:datasets}. Some implementation details are depicted in Section~\ref{subsec:implementationdetails}, followed by curved text detection results in Section~\ref{subsec:curvetextresults}. The experimental results on multi-oriented text detection is given in Section~\ref{subsec:straighttextresults} to demonstrate the versatility of the proposed TextField. To further demonstrate the generality of TextField, cross dataset experiments are also presented in Section~\ref{subsec:crossdataset}. The runtime analysis and some failures cases are given in Section~\ref{subsec:runtime} and Section~\ref{subsec:Weakness}, respectively.

\subsection{Datasets and evaluation protocol}
\label{subsec:datasets}
\noindent\textbf{SynthText in the Wild}~\cite{gupta2016synthetic}: SynthText contains 800k synthetic images generated by blending natural images with artificial text. Annotations are given in character, word, and line level. This dataset with word level annotation is used to pre-train the proposed model.

\medskip

\noindent\textbf{SCUT-CTW1500}~\cite{liu2017detecting}: Different from classical multi-oriented text datasets, this dataset is quite challenging due to many curved texts. It consists of 1000 training images and 500 testing images. This dataset has more than 10k text annotations and at least one curved text per image. Each text instance is labeled by a polygon with 14 points. The annotation is given in line or curve level.

\medskip

\noindent\textbf{Total-Text}~\cite{ch2017total}: Total-Text dataset also aims at solving the arbitrary-shaped text detection problem. It contains 1555 scene images, divided into 1255 training images and 300 testing images. This dataset contains many curved and multi-oriented texts. Annotations are given in word level with polygon-shaped bounding boxes instead of conventional rectangular bounding boxes.

\medskip

\noindent\textbf{ICDAR2015 Incidental Scene Text (IC15)}~\cite{karatzas2015icdar}: This dataset is widely used to benchmark multi-oriented text detectors. It was released for the Challenge 4 of ICDAR2015 Robust Reading Competition. Different from previous datasets with text captured in relatively high resolution, scene images in this dataset are taken by Google Glasses in an incidental manner. Therefore, text in these images is of various scales, orientations, contrast, blurring, and viewpoint, making it challenging for detection. This dataset is composed of 1000 training images and 500 testing images. Annotations are provided with word-level bounding quadrilaterals.

\medskip

\noindent\textbf{MSRA-TD500}~\cite{yao2012detecting}: This dataset is dedicated for detecting multi-lingual long texts of arbitrary orientations. It consists of 300 training images and 200 testing images, annotated at the level of text lines. Since the number of training images is rather small, similar with other methods, we also utilize the images from HUST-TR400~\cite{yao2014unified} as extra training data.

\medskip

\noindent\textbf{Evaluation protocol}: We follow the standard evaluation protocol relying on $precision$, $recall$, and $f$-$measure$. Precisely, they are defined as following:
\begin{align}
\begin{split}
precision &= \frac{TP}{TP+FP},\\
Recall &= \frac{TP}{TP+FN},\\
f\textrm{-}measure &= 2 \times \frac{precision \times recall}{precision + recall},
\end{split}
\label{eq:measure}
\end{align}
where $TP$, $FP$, and $FN$ stands for the number of correctly detected text instances, incorrect detections, and missing text instances, respectively. For a detected text instance $T$, if $T$ intersects a ground truth text instance with an IOU larger than a given thresholding value (typically set to 0.5), then the text instance $T$ is considered as a correct detection. Since there is a trade-off between $recall$ and $precision$, $f\textrm{-}measure$ is a common compromised measurement for performance assessment.

\subsection{Implementation Details}
\label{subsec:implementationdetails}
Data augmentation strategy is adopted to increase the training data and avoid over-fitting. Specifically, images are first randomly cropped with area ratios ranging from 0.1 to 1 and aspect ratios ranging from 0.3 to 3. The cropped image is then randomly rotated with $0$ and $\pm 90$ degrees. Note that the randomly cropped patch is selected only when the proportion of contained texts with respect to all ground truth text areas in the original images is larger than a threshold value, randomly set to 0.1, 0.3, 0.5, and 0.7. Finally, the augmented images are resized to $384 \times 384$ or $768 \times 768$ during different training stages detailed in the following.

The proposed network is pre-trained on SynthText for one epoch, and then finetuned on SCUT-CTW1500, Total-Text, ICDAR2015 Incidental Scene Text, and MSRA-TD500, respectively. The training process is divided into three stages. In the pre-training stage, the augmented images are resized to $384 \times 384$ for the sake of training speed. The learning rate and the hyper-parameter $\gamma$ involved in online hard negative mining are set to $10^{-4}$ and 3, respectively. Then we finetune our model on each dataset for about 100 epochs with the same settings as pre-training stage. We continue to train the network for another 100 epochs by resizing the augmented images to $768 \times 768$ aiming at better handling multi-scale texts. In this last training stage, the learning rate is decayed to $10^{-5}$ and $\gamma$ is set to 6. In the whole training process, we adopt Adam~\cite{kinga2015method} to optimize the network. All the experiments are conducted on Caffe~\cite{jia2014caffe} using a workstation with a single Titan Xp GPU.

\subsection{Curved text detection}
\label{subsec:curvetextresults}
The proposed TextField is appropriate to detect irregular texts. We first conduct experiments on two curved text datasets: SCUT-CTW1500 and Total-Text.

\textbf{SCUT-CTW1500}:
This dataset mainly contains curved and multi-oriented texts. For each image, the annotation is given in line or curve level. The size of testing image is rather small. In testing phase, the images are resized to $576 \times 576$. The threshold parameter $\lambda_m$ is set to 0.59 for post-processing. A visualization example of the learned direction field and some involved post-processing steps is depicted in Fig.~\ref{fig:postprocessing_eg}(a). Some qualitative results are given in Fig.~\ref{fig:qualitativeresults}(a). The proposed TextField correctly detects text of arbitrary shapes with very accurate text boundaries. The quantitative results are shown in Tab.~\ref{tab:ctw1500quantitativeresults}. Compared with other state-of-the-art methods, our proposed method outperforms them by a large margin in terms of recall, precision, and f-measure. The proposed TextField achieves 81.4\% F-measure, improving the state-of-the-art methods by 8.0\%.

\begin{table}[!tbp]
\centering
\caption{Quantitative results of different methods evaluated on SCUT-CTW1500. $^*$ indicates the result obtained from~\cite{liu2017detecting}.}
\label{tab:ctw1500quantitativeresults}
\begin{tabular}{|c|c|c|c|}
\hline
Methods      & recall  & precision  & f-measure \\ \hline
SegLink $^*$~\cite{shi2017seglink}      & 0.400   & 0.423      & 0.408      \\ \hline
CTPN $^*$~\cite{tian2016ctpn}         & 0.538   & 0.604      & 0.569      \\ \hline
EAST $^*$~\cite{zhou2017east}         & 0.491   & 0.787      & 0.604      \\ \hline
DMPNet $^*$~\cite{liu2017deep}       & 0.560   & 0.699      & 0.622      \\ \hline
CTD~\cite{liu2017detecting}          & 0.652   & 0.743      & 0.695      \\ \hline
CTD+TLOC~\cite{liu2017detecting}     & 0.698   & 0.774      & 0.734      \\ \hline
TextField (Ours)         & \bf{0.798}  & \bf{0.830}  & \bf{0.814}  \\ \hline
\end{tabular}
\end{table}

\begin{table}[!tbp]
\centering
\caption{Quantitative results of different methods evaluated on Total-Text.}
\label{tab:totaltextquantitativeresults}
\begin{tabular}{|c|c|c|c|}
\hline
Methods      & recall  & precision  & f-measure \\ \hline
Ch'ng $et$ $al.$~\cite{ch2017total}       & 0.400   & 0.330      & 0.360      \\ \hline
Liao $et$ $al.$~\cite{liao2017textboxes}         & 0.455   & 0.621      & 0.525      \\ \hline
TextField (Ours)         & \bf{0.799}  & \bf{0.812}  & \bf{0.806}  \\ \hline
\end{tabular}
\end{table}

\textbf{Total-Text}:
We also evaluate the proposed TextField on Total-Text whose annotations are given in word level. This dataset mainly contains curved and multi-oriented texts. In testing, all images are resized to $768 \times 768$. The threshold parameter $\lambda_m$ is set to 0.50 for post-processing. A visualization example of the learned direction field and some involved post-processing steps is illustrated in Fig.~\ref{fig:postprocessing_eg}(b). Some qualitative results are depicted in Fig.~\ref{fig:qualitativeresults}(b). From this figure, we can observe that TextField also precisely detects word level irregular texts. And TextField is able to accurately separate close text instances of arbitrary shapes. The quantitative results are given in Tab.~\ref{tab:totaltextquantitativeresults}. The proposed TextField achieves 80.6\% F-measure on this dataset, significantly outperforming other methods.

From the qualitative results depicted in Fig.~\ref{fig:qualitativeresults}(a-b) and quantitative results given in Tab.~\ref{tab:ctw1500quantitativeresults} and Tab.~\ref{tab:totaltextquantitativeresults}, the proposed TextField is able to detect irregular texts in both line-level and word-level. TextField establishes new state-of-the-art results in detecting curved texts.

\subsection{Multi-oriented text detection}
\label{subsec:straighttextresults}
As shown in Section~\ref{subsec:curvetextresults}, the proposed TextField significantly outperforms other methods on curved text detection. To further demonstrate the ability of TextField in detecting texts of arbitrary shapes, we evaluate TextField on ICDAR2015 Incidental Scene Text and MSRA-TD500 dataset, showing that TextField also achieves very competitive results on widely adopted multi-oriented datasets. Note that for these two experiments, we fit each text instance achieved with TextField by a minimum oriented bounding rectangle.

\begin{figure*}
\centering
\begin{subfigure}[b]{1.0\textwidth}
\centering
\includegraphics[width=1.0\linewidth]{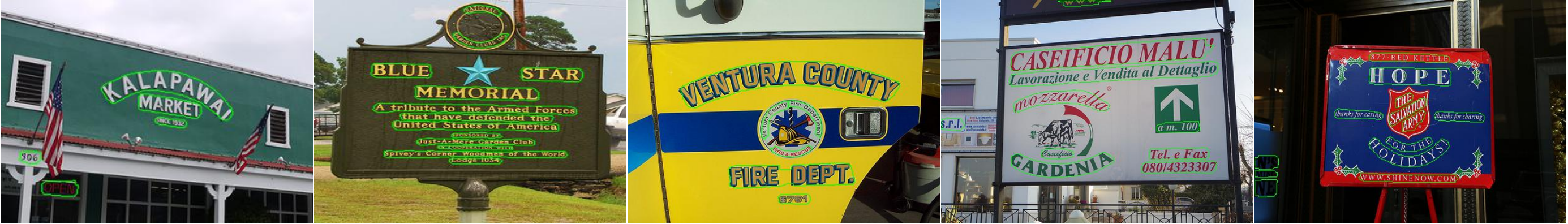}
\caption{}
\label{fig:ctwresults}
\end{subfigure}

\vspace{1mm}
\begin{subfigure}[b]{1.0\textwidth}
\centering
\includegraphics[width=1.0\linewidth]{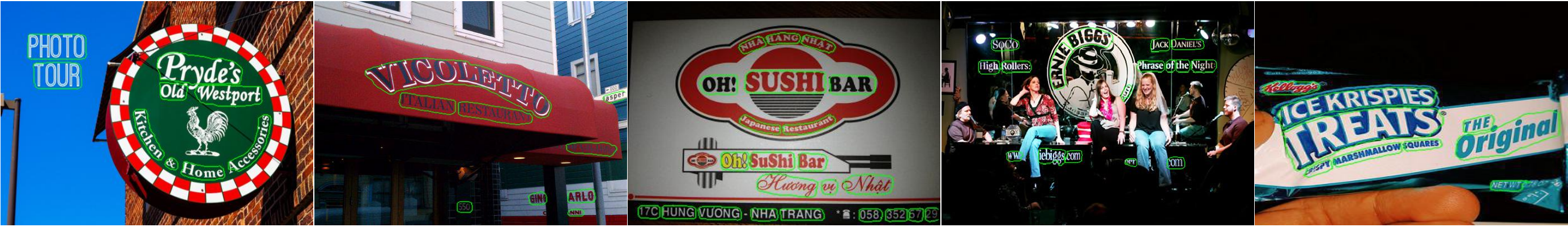}
\caption{}
\label{fig:totaltextresults}
\end{subfigure}

\vspace{1mm}
\begin{subfigure}[b]{1.0\textwidth}
\centering
\includegraphics[width=1.0\linewidth]{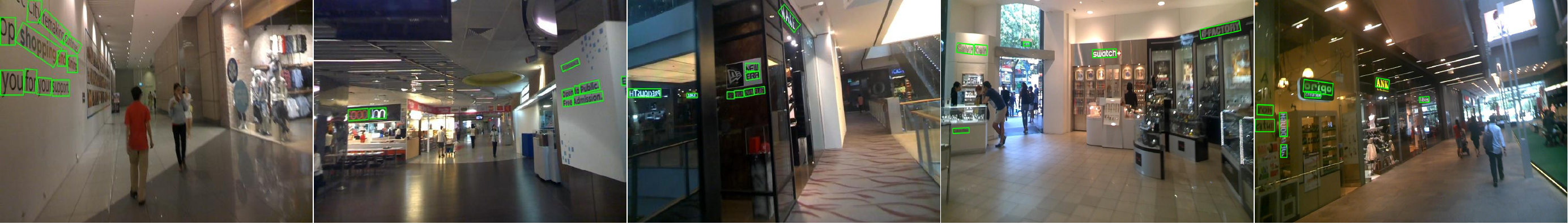}
\caption{}
\label{fig:ic15results}
\end{subfigure}

\vspace{1mm}
\begin{subfigure}[b]{1.0\textwidth}
\centering
\includegraphics[width=1.0\linewidth]{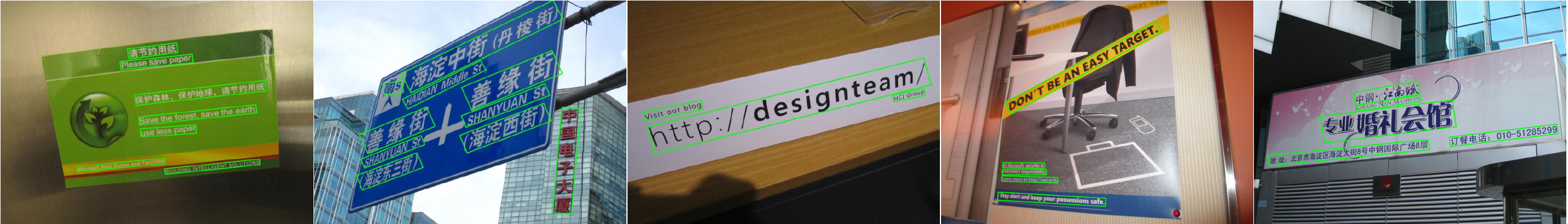}
\caption{}
\label{fig:td500results}
\end{subfigure}
\caption{Some qualitative detection results on SCUT-CTW500 in (a), Total-Text in (b), IC15 in (c), and MSRA-TD500 in (d). The arbitrary-shaped texts are correctly detected with accurate text instance boundaries.}
\label{fig:qualitativeresults}
\end{figure*}

\textbf{ICDAR2015 Incidental Scene Text}:
Images in this dataset are of low resolution and contain many small text instances. Therefore, images are not resized. The original resolution of $1280 \times 720$ is used in testing. The threshold parameter $\lambda_m$ is set to 0.69 for post-processing. A visualization example of the learned direction field and some involved post-processing steps is shown in Fig.~\ref{fig:postprocessing_eg}(c).
Some detection results on this dataset are given in Fig.~\ref{fig:qualitativeresults}(c), where challenging texts of variant contrast and scales are correctly detected. The quantitative evaluation compared with other methods are depicted in Tab.~\ref{tab:ic15quantitativeresults}. The proposed TextField achieves competitive results with other state-of-the-art methods on this dataset. Following ~\cite{he2017DDR,zhou2017east,liao2018textboxes++,lyu2018multi}, we also report the results of TextField under multi-scale evluation using $384 \times 384$, $768 \times 768$, and $1024 \times 1024$ inputs on IC15. TextField is also very competitive with other methods under multi-scale evaluation. Note that for fair comparison, we mainly compare with other methods using the same backbone network (\ie, VGG16 network).

\begin{table}[!tbp]
\centering
\caption{Comparison of methods on ICDAR2015 Incidental Scene Text. $^\dagger$ means that the base net of the model is not VGG16. $^*$ stands for multi-scale version.}
\label{tab:ic15quantitativeresults}
\begin{tabular}{|c|c|c|c|c|}
\hline
Methods      & recall  & precision  & f-measure  & FPS\\ \hline
Zhang $et$ $al.$~\cite{zhang2016textfcn}         & 0.430   & 0.708      & 0.536   & 0.48   \\ \hline
CTPN~\cite{tian2016ctpn}         & 0.516   & 0.742      & 0.609   & 7.1   \\ \hline
Yao $et$ $al.$~\cite{yao2016scene}         & 0.587   & 0.723      & 0.648   & 1.61   \\ \hline
DMPNet~\cite{liu2017deep} & 0.682 & 0.732 & 0.706 & - \\
\hline
SegLink~\cite{shi2017seglink}      & 0.768   & 0.731      & 0.750   & -   \\ \hline
MCN~\cite{liu2018learning} & 0.800 & 0.720 & 0.760 & - \\
\hline
EAST~\cite{zhou2017east}         & 0.728   & 0.805      & 0.764   & 6.52   \\ \hline
SSTD~\cite{he2017single}         & 0.730   & 0.800      & 0.770   & 7.7   \\ \hline
RRPN~\cite{ma2018arbitrary} & 0.730 & 0.820 & 0.770 & - \\
\hline
ITN~\cite{wang2018geometry} & 0.741 & 0.857 & 0.795 & -
\\ \hline
EAST $^{\dagger}$~\cite{zhou2017east}         & 0.735   & 0.836      & 0.782   & 13.2   \\ \hline
Lyu $et$ $al.$~\cite{lyu2018multi}         & 0.707   & \bf{0.941}      & 0.807   & 3.6   \\ \hline
TextBoxes++~\cite{liao2018textboxes++}         & 0.767   & 0.872      & 0.817   & 11.6   \\ \hline
RRD~\cite{liao2018rotation} & 0.790 & 0.856 & 0.822 & 6.5 \\
\hline
TextField (Ours)         & \textbf{0.805}  &  {0.843}  & \textbf{0.824}  & 6.0 \\ \hline\hline
WordSup $^*$~\cite{hu2017wordsup} & 0.770   & 0.793 & 0.782 & 2
\\ \hline
EAST $^{\dagger *}$~\cite{zhou2017east} & 0.783 & 0.833 & 0.807 & -
\\ \hline
He $et$ $al.$ $^{\dagger *}$~\cite{he2017DDR} & 0.800   & 0.820 & 0.810 & 1.1
\\ \hline
TextBoxes++ $^*$~\cite{liao2018textboxes++} & 0.785 & 0.878 & 0.829 & 2.3
\\ \hline
Lyu $et$ $al.$ $^*$~\cite{lyu2018multi} & 0.797 & \bf{0.895} & \bf{0.843} & 1
\\ \hline
TextField $^*$ (Ours) & \bf{0.839} & 0.843 & 0.841 & 1.8
\\ \hline
\end{tabular}
\end{table}

\textbf{MSRA-TD500}:
This dataset contains both English and Chinese texts whose annotations are given in terms of text lines. The text scale varies significantly. In testing, we resize the images into $768 \times 768$. The threshold parameter $\lambda_m$ is set to 0.64 for post-processing. Due to the large character spacing in this dataset, we also group the detected texts with small aspect ratios before evaluating the TextField using the IC15 evaluation code. A visualization example of the learned direction field and some involved post-processing steps is depicted in Fig.~\ref{fig:postprocessing_eg}(d). Some qualitative illustrations are shown in Fig.~\ref{fig:qualitativeresults}(d). The proposed TextField successfully detects long text lines of arbitrary orientations and sizes. The quantitative comparison with other methods on this dataset is given in Tab.~\ref{tab:td500quantitativeresults}. TextField also achieves competitive performance with other methods in detecting long multi-oriented texts. Specifically, TextField performs slightly worse than the methods in~\cite{lyu2018multi} and~\cite{liu2018learning} on MSRA-TD500. Yet, the performance of TextField is much better than them on IC15 dataset.

From the qualitative results in Fig.~\ref{fig:qualitativeresults}(c-d) and quantitative evaluations in Tab.~\ref{tab:ic15quantitativeresults} and Tab.~\ref{tab:td500quantitativeresults}, the proposed TextField is also capable to accurately detect multi-oriented texts in both line-level and word-level. This demonstrates the versatility of the proposed TextField.

\begin{table}[!tbp]
\centering
\caption{Comparison of methods on MSRA-TD500. $^\dagger$ stands for the base net of the model is not VGG16.}
\label{tab:td500quantitativeresults}
\begin{tabular}{|c|c|c|c|}
\hline
Methods      & recall  & precision  & f-measure
\\ \hline
He $et$ $al.$~\cite{he2016text} & 0.610 & 0.760 & 0.690
\\ \hline
EAST~\cite{zhou2017east}         & 0.616   & 0.817      & 0.702 \\
\hline
ITN~\cite{wang2018geometry} & 0.656 & 0.803 & 0.722 \\
\hline
Zhang $et$ $al.$~\cite{zhang2016textfcn}         & 0.670   & 0.830      & 0.740
\\ \hline
RRPN~\cite{ma2018arbitrary} & 0.680 & 0.820 & 0.740
\\ \hline
He $et$ $al.$ $^{\dagger}$~\cite{he2017DDR}         & 0.700   & 0.770      & 0.740
\\ \hline
Yao $et$ $al.$~\cite{yao2016scene}         & 0.753   & 0.765      & 0.759
\\ \hline
EAST $^{\dagger}$~\cite{zhou2017east}         & 0.674   & 0.873      & 0.761
\\ \hline
Wu $et$ $al.$~\cite{wu2017self} & 0.780 & 0.770 & 0.770 \\
\hline
SegLink~\cite{shi2017seglink}      & 0.700   & 0.860      & 0.770
\\ \hline
RRD~\cite{liao2018rotation} & 0.730 & 0.870 & 0.790
\\ \hline
Lyu $et$ $al.$~\cite{lyu2018multi}         & 0.762   & 0.876      & 0.815
\\ \hline
MCN~\cite{liu2018learning} & \bf{0.790} & \bf{0.880} & \bf{0.830}
\\ \hline
TextField (Ours)         & {0.759}  &  {0.874}  & {0.813}
\\ \hline
\end{tabular}
\end{table}

\subsection{Cross dataset text detection}
\label{subsec:crossdataset}

\begin{table}[tbp]
\centering
\caption{Cross-dataset evaluations of different methods on corresponding word-level and line-level datasets.}
\label{tab:crossdatasetresults}
\begin{tabular}{|c|c|c|c|}
\hline
\multirow{2}{*}{Methods}
& \multicolumn{3}{c|}{Total-Text (train on IC15)}
\\ \cline{2-4}
& recall     & precision       & f-measure
\\ \hline
SegLink~\cite{shi2017seglink} & 0.332 & 0.356 & 0.344
\\ \hline
EAST~\cite{zhou2017east} & 0.431 & 0.490 & 0.459
\\ \hline
TextField (Ours)
& \bf{0.652}  & \bf{0.615}  & \bf{0.633}
\\ \hline
\multirow{2}{*}{Methods}
& \multicolumn{3}{c|}{IC15 (train on Total-Text)}
\\ \cline{2-4}
& recall     & precision       & f-measure
\\ \hline
TextField (Ours) & \bf{0.660}  & \bf{0.771}  & \bf{0.711}
\\ \hline
\multirow{2}{*}{Methods}
& \multicolumn{3}{c|}{SCUT-CTW1500 (train on TD500)}
\\ \cline{2-4}
& recall     & precision       & f-measure
\\ \hline
TextField (Ours) & \bf{0.700}  & \bf{0.753}  & \bf{0.726}
\\ \hline
\multirow{2}{*}{Methods}
& \multicolumn{3}{c|}{MSRA-TD500 (train on SCUT-CTW1500)}
\\ \cline{2-4}
& recall     & precision       & f-measure
\\ \hline
TextField (Ours) & \bf{0.758}  & \bf{0.853}  & \bf{0.803}
\\ \hline
\end{tabular}
\end{table}

To further demonstrate the generalization ability of the proposed TextField, we also evaluate the TextField trained on one dataset and test the trained model on a different dataset annotated in the same level (\eg, word or line). Specifically, we first benchmark several classical models (trained on IC15) on Total-Text dataset. As depicted in Tab.~\ref{tab:crossdatasetresults}, The proposed TextField generalizes better on cross-dataset text detection. We then test the TextField (trained on Total-Text) on IC15 dataset, which gives acceptable results. We have also performed cross-dataset evaluations on two line-level annotated datasets: SCUT-CTW1500 and MSRA-TD500. As shown in Tab.~\ref{tab:crossdatasetresults}, for the line-based text detection, TextField also achieves very competitive results (under cross-dataset setting) with some state-of-the-art methods trained on the target dataset. Specifically, TextField trained on MSRA-TD500 containing only multi-oriented texts performs comparably with other methods properly trained on SCUT-CTW1500, a curved text dataset. Furthermore, it is worth to note that TextField trained on SCUT-CTW1500 containing mainly curved English texts also performs rather well (with a small degradation) in detecting multi-oriented Chinese texts in MSRA-TD500.

These cross-dataset experiments demonstrate that the proposed TextField is effective in detecting irregular texts, and is also robust in generalizing to unseen datasets.

\subsection{Runtime}
\label{subsec:runtime}
The proposed TextField first yields the predicted direction field through the proposed network, then followed by a morphological post-processing step to achieve final text detection results. The runtime of TextField is thus decomposed into two stages: network inference and post-processing. For the network inference, using the VGG16 backbone network as depicted in Fig.~\ref{fig:networkarchitecture}, it takes about 130ms for a $1280 \times 720$ IC15 image and 100ms for a $768 \times 768$ MSRA-TD500 image on a Titan Xp GPU. As described in Section~\ref{subsec:postprocessing}, the post-processing is mainly composed of three steps: text superpixel segmentation, text superpixel grouping, and text instance filtering. The text superpixel segmentation could be achieved by the blob labeling algorithm which is very fast. The grouping step only involves some classical morphological operations. The text instance filtering step is also very fast thanks to the criterion incrementally computed during the grouping step. The whole post-processing stage takes about 36ms for a $1280 \times 720$ IC15 image and 24ms for a $768 \times 768$ MSRA-TD500 image on a 3.4GHz/8MB cache Intel core i7-2600, 16GB RAM. As depicted in Tab.~\ref{tab:ic15quantitativeresults}, the proposed TextField runs at 6.0 FPS using VGG16 backbone, which is on par with most state-of-the-art methods. Furthermore, TextField is able to accurately detect irregular texts and generalizes well to unseen datasets.

\subsection{Weakness}
\label{subsec:Weakness}
\begin{figure}
\centering
\includegraphics[width=1.0\linewidth]{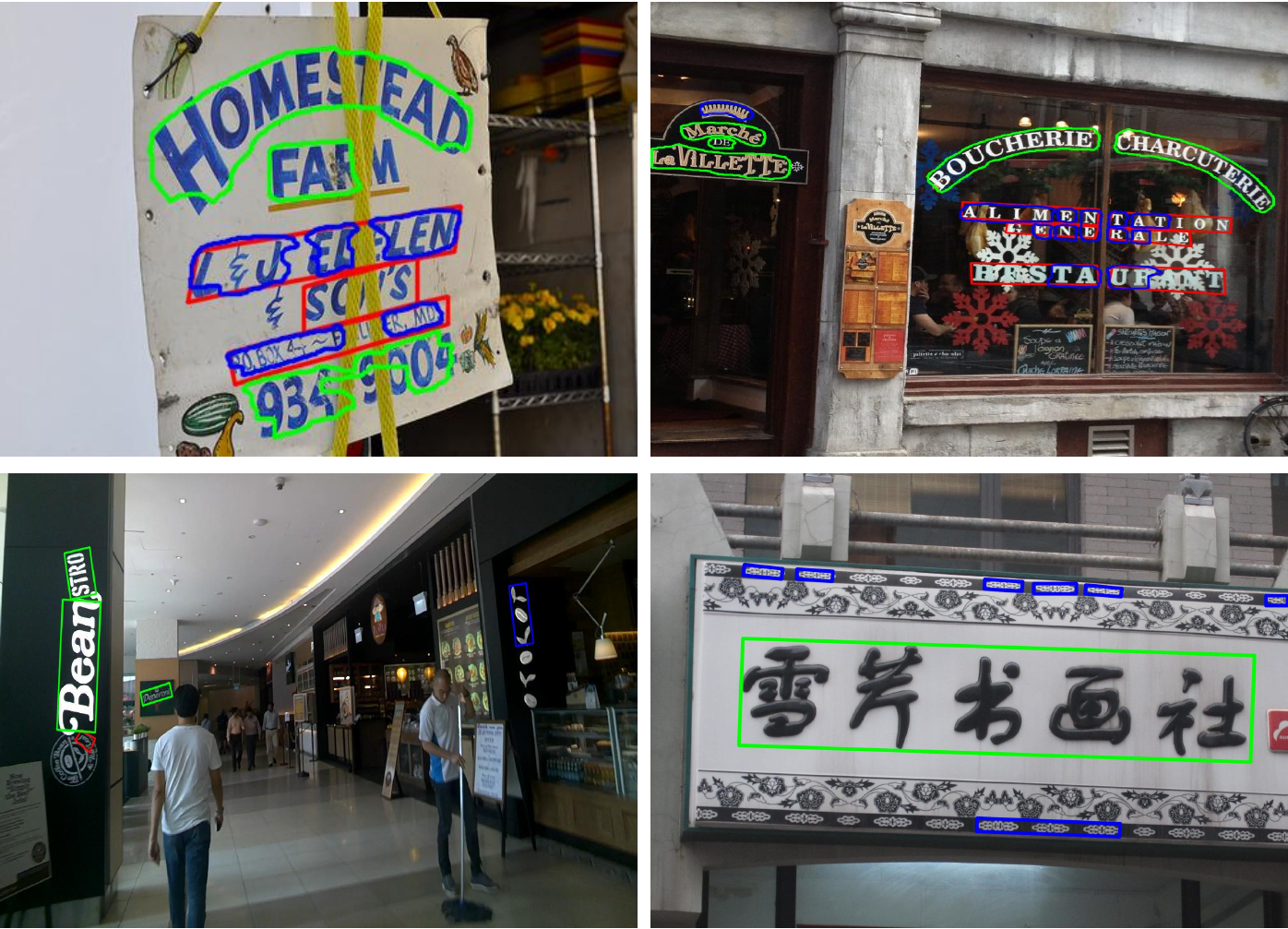}
\caption{Some failure examples. Green contours: correct detections; Red contours: missing ground truths; Blue contours: false detections.}
\label{fig:failures}
\end{figure}
As demonstrated in previous experiments, TextField performs well in most cases of detecting texts of arbitrary shapes. It still fails for some difficult images, such as object occlusion, large character spacing. Some failure examples are given in Fig.~\ref{fig:failures}. TextField also has some false detections on some text-like areas. Note that all these difficulties are common challenges for the other state-of-the-art methods~\cite{zhou2017east,shi2017seglink,liao2018textboxes++}.

\section{Conclusion}\label{sec:conclusion}

We have presented TextField, which learns a deep direction field for irregular text detection. Specifically, we propose a novel text direction field that points away from nearest text boundary to each text point. Such two-dimensional text direction field encodes both binary text mask and direction information that facilitates the separation of adjacent text instances, which remains challenging for classical segmentation-based approaches. TextField directly regresses the direction field followed by a simple yet effective post-processing step inspired by some morphological tools. Experiments on two curved text datasets (Total-Text and SCUT-CTW1500) and two widely-used datasets (ICDAR 2015 and MSRA-TD500) demonstrate that the proposed method outperforms all state-of-the-art methods by a large margin in detecting curved texts, and achieves very competitive performances in detecting multi-oriented texts. Furthermore, based on the cross-dataset evaluations, TextField also generalizes well to unseen datasets. In the future, we would explore more robust text superpixel grouping strategy ({\em e.g.,} via explicitly learning the text center line) to further boost TextField, and investigate the common challenges faced by all state-of-the-art text detectors.


\bibliographystyle{IEEEtran}

\bibliography{main}

\end{document}